\documentclass[11pt]{article}

\usepackage[final]{acl}

\usepackage{amsfonts}
\usepackage{subcaption}
\usepackage{amsmath}

\usepackage{times}
\usepackage{latexsym}
\usepackage{wrapfig}
\usepackage{multirow}
\usepackage{booktabs}
\usepackage[T1]{fontenc}

\usepackage[T1]{fontenc}

\usepackage[utf8]{inputenc}

\usepackage{microtype}

\usepackage{inconsolata}

\usepackage{hyperref}
\usepackage{url}
\usepackage{graphicx}

\title{TIGER: Text-Informed Generalized Enzyme-Reaction Retrieval}

\author{
 \textbf{Yuhang Zhang\textsuperscript{1,2}},
 \textbf{Keyan Ding\textsuperscript{3}},
 \textbf{Peilin Chen\textsuperscript{2}},
 \textbf{Han Liu\textsuperscript{2}},
\\
 \textbf{Can Lin\textsuperscript{1}},
 \textbf{Ruixi Chen\textsuperscript{1}},
 \textbf{Shiqi Wang\textsuperscript{2}\thanks{Corresponding authors.}},
 \textbf{Qi Song\textsuperscript{1}\footnotemark[1]}
\\
\\
 \textsuperscript{1}University of Science and Technology of China\\
 \textsuperscript{2}City University of Hong Kong~~
 \textsuperscript{3}Zhejiang University
\\
 \\
\texttt{
   \href{mailto:yhzhang@mail.ustc.edu.cn}{yhzhang}@mail.ustc.edu.cn,
   \href{mailto:dingkeyan@zju.edu.cn}{dingkeyan}@zju.edu.cn,
   \href{mailto:plchen3@cityu.edu.hk}{plchen3}@cityu.edu.hk,
   }\\
\texttt{
   \href{mailto:hanliu.sdu@gmail.com}{hanliu.sdu}@gmail.com,
   \{\href{mailto:can.lin@mail.ustc.edu.cn}{can.lin}, \href{mailto:ruixichen@mail.ustc.edu.cn}{ruixichen}\}@mail.ustc.edu.cn,
   }\\
\texttt{
   \href{mailto:shiqwang@cityu.edu.hk}{shiqwang}@cityu.edu.hk, \href{mailto:qisong09@ustc.edu.cn}{qisong09}@ustc.edu.cn
 }
}

\begin{document}
\maketitle
\begin{abstract}
Enzyme–Reaction Retrieval is a fundamental problem in computational biology, underpinning enzyme characterization, reaction mechanism elucidation, and the rational design of metabolic pathways and biocatalysts. As a bidirectional task, it entails both enzyme-to-reaction and reaction-to-enzyme mapping. However, existing approaches suffer from poor generalization across tasks and distributions, with performance highly sensitive to dataset splits and substantial asymmetry between retrieval directions. To address these challenges, we present \textit{TIGER, a Text-Informed Generalized Enzyme-Reaction Retrieval framework} that leverages protein-to-text generation models to distill textual semantic knowledge from enzyme sequences, providing a generalized representation that bridges enzymes and biochemical reactions. To ensure the quality and reliability of textual semantics, we design a \textit{Dynamic Gating Network} that adaptively fuses text-derived knowledge with sequence features, enabling more consistent and informative enzyme representations, while a \textit{Structure-Shared Feature Projector} aligns enzyme and reaction representations within a unified latent space. Extensive experiments demonstrate that, under bidirectional retrieval supervision, TIGER significantly outperforms state-of-the-art baselines across diverse distributions and exhibits strong robustness and transferability across tasks.
\end{abstract}

\begin{figure*}[h]
    \centering
    \includegraphics[width=\linewidth]{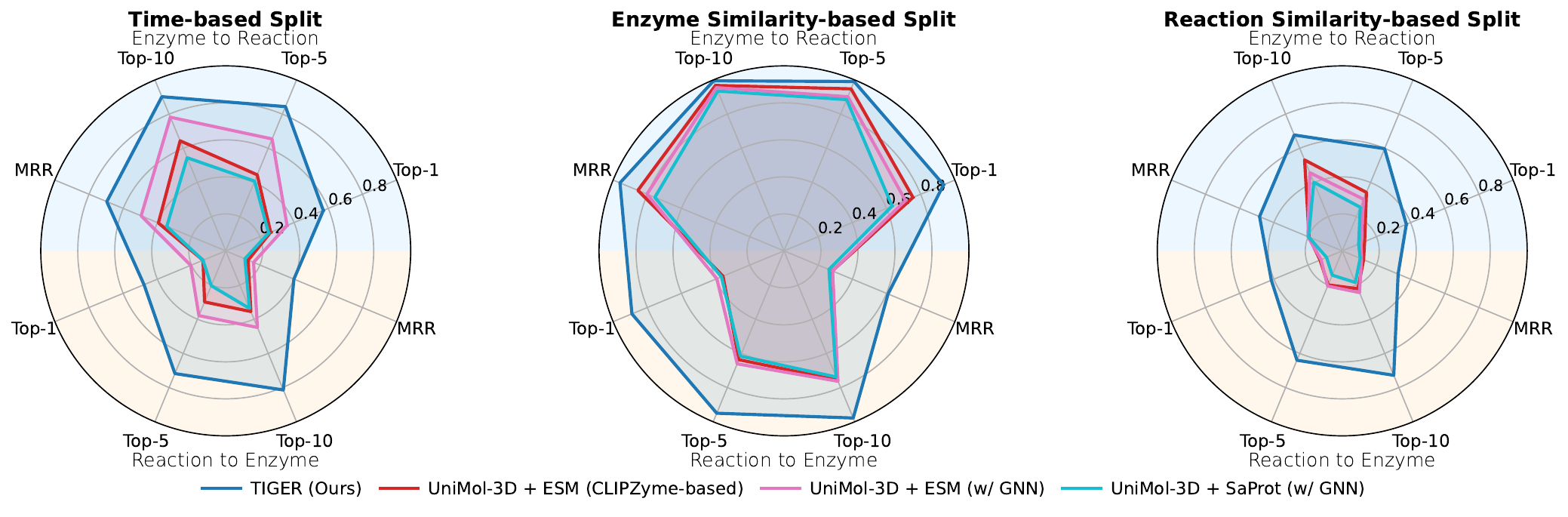}
    \caption{Bidirectional Retrieval performance of TIGER and existing methods under time-, enzyme similarity-, and reaction similarity-based splits on ReactZyme, demonstrating the robust generalization capacity of TIGER across heterogeneous evaluation settings.}
    \label{fig:Spider}
\end{figure*}

\section{Introduction}
Enzymes~\cite{buller2023nature,benitez2022multistep}, as central biocatalysts, orchestrate biochemical transformations essential for life. While traditional bioinformatics has focused on predictive tasks such as EC number classification~\cite{yu2023enzyme,dalkiran2018ecpred}, the emerging field of Enzyme–Reaction Retrieval adopts a relational paradigm that models bidirectional correspondences between enzymes and their catalyzed reactions. This enables systematic exploration of functional diversity, pathway reconstruction, and synthetic biology applications. With the rapid growth of high-throughput sequencing~\cite{pai2021high,hoffman2014high}, the expanding scale of enzyme–reaction data offers fertile ground for such modeling advances.

Existing computational approaches for enzyme-reaction retrieval have predominantly relied on contrastive learning paradigms~\cite{mikhael2024clipzyme}, which align representations derived from enzyme sequences with those from chemical reactions. However, these frameworks exhibit notable limitations that hinder their practical application~\cite{hua2024reactzyme}. First, \textit{they demonstrate bidirectional asymmetry}, where the retrieval accuracy from enzymes to reactions substantially diverges from the reverse direction. This asymmetry reveals a fundamental inconsistency in semantic alignment and a lack of representational coherence. Second, \textit{these models show a high sensitivity to dataset splits}, with performance fluctuating significantly under different partition strategies. This instability points to a critical lack of generalization ability and raises concerns about their robustness across heterogeneous data distributions.

A key reason for the performance gap is that pre-trained protein models are not explicitly trained to understand chemical transformations. Their objectives emphasize structural and evolutionary signals~\cite{lin2023esm2,jumper2021alphafold,wang2022single}, rather than the reaction-specific features required for catalysis. Motivated by recent advances in knowledge-enhanced multimodal retrieval tasks~\cite{mi2024knowledge,feng2023mkvse,suo2024knowledge}, we propose Text-Informed Generalized Enzyme–Reaction Retrieval (TIGER) framework, which leverages protein-to-text generation models~\cite{liu2024prott,abdine2024prot2text} to produce rich, knowledge-aware enzyme descriptions. These textual summaries move beyond sequential or structural signals by encoding catalytic function, substrate interactions, and other reaction-linked properties, enabling a more symmetric and semantically aligned joint embedding space.

As textual descriptions generated by pre-trained language models are prone to semantic noise~\citep{cao2025survey,liang2024controllable} and ``hallucinations''~\citep{vishwanath2024faithfulness,jesson2024estimating}, we introduce a Dynamic Gating Network (DGN) to adaptively regulate their contribution during representation learning. Instead of treating all textual inputs equally, the DGN learns reliability-aware gating weights that reflect the semantic consistency of textual embeddings with enzyme sequence features. Reliable descriptions are thus emphasized to enrich biochemical semantics, while noisy or irrelevant ones are down-weighted to prevent spurious correlations. This adaptive modulation enables the framework to retain the complementary knowledge conveyed by textual cues while enhancing robustness against imperfect supervision, ultimately leading to more stable training and stronger generalization in enzyme–reaction retrieval. 

To enhance representational coherence and cross-modal generalization, we introduce a Structure-Shared Feature Projector that maps enzyme and reaction embeddings into a unified latent space. We trained TIGER with bidirectional contrastive supervision and evaluated it on ReactZyme, the largest enzyme–reaction dataset available. As shown in Figure~\ref{fig:Spider}, TIGER consistently outperforms representative baselines across time-based, enzyme similarity-based, and reaction similarity-based splits, demonstrating superior retrieval accuracy and robustness. These results highlight its strong generalization ability under diverse conditions and underscore the effectiveness of the text-informed design. In summary, our main contributions are:
\begin{itemize}
    \item We propose\textit{ TIGER, a text-informed generalized enzyme–reaction retrieval framework} that incorporates knowledge-rich descriptions to establish a more symmetric and semantically coherent embedding space.
    \item A \textit{Dynamic Gating Network} is introduced to robustly balance sequence and text signals under noisy AI-generated descriptions, together with a \textit{Structure-Shared Feature Projector} that aligns both modalities in a unified embedding space to enhance cross-modal generalization.
    \item We conduct comprehensive experiments on ReactZyme, where TIGER consistently achieves state-of-the-art performance, yielding relative Hit@1 improvements ranging \textit{from 14\% to over 200\%} across diverse evaluation splits, with ablation studies confirming the contribution of each component.
\end{itemize}

\section{Related Work}
\textbf{Enzyme-Reaction Retrieval}\footnote{Task formulation is introduced in the Appendix(\ref{sec:task}).}
Traditional enzyme studies have largely focused on EC classification~\cite{dalkiran2018ecpred,yu2023enzyme} and substrate binding~\cite{zeng2022substrate}, yet such categorical tasks~\cite{fernstad2011task} overlook the richer relational structure between enzymes and the reactions they catalyze. Enzyme-reaction retrieval has thus emerged as a more flexible paradigm, learning a shared embedding space to directly align and rank enzymes with reactions.
Early attempts such as CLIPZyme~\cite{mikhael2024clipzyme} leveraged contrastive learning to couple enzymatic sequences with reaction representations, offering a relatively initial yet advanced formulation of enzyme–reaction retrieval. To further advance this emerging direction, ReactZyme~\cite{hua2024reactzyme} established a large-scale standardized benchmark that integrates diverse enzyme–reaction data and enables systematic evaluation across multiple methodologies. In particular, it benchmarked a spectrum of baselines, including 2D and 3D molecular encoders for reactions (MAT-2D/3D~\cite{maziarka2020molecule}, UniMol-2D/3D~\cite{zhou2023uni}), protein language models for enzymes (ESM~\cite{lin2023esm2}, SaProt~\cite{su2023saprot}), and residue-level equivariant graph networks (FANN)~\cite{puny2021frame}, thereby providing a comprehensive testbed for cross-modal retrieval.
Importantly, these experiments further revealed open challenges, including bidirectional asymmetry and sensitivity to dataset splits, underscoring the need for more robust and semantically grounded retrieval frameworks.

\textbf{Text-informed Protein Representation Learning}
Protein representation learning has traditionally focused on amino acid sequences, yet such unimodal formulations overlook the rich functional and mechanistic semantics available in biomedical corpora. Recent progress has therefore moved toward text-informed paradigms, where protein sequences are complemented with natural language supervision to obtain more expressive, function-aware embeddings.
Early work such as ProtST~\cite{xu2023protst} aligned protein sequences with biomedical text using dual-modal contrastive learning, demonstrating that textual context provides orthogonal signals that enhance downstream tasks. ProTrek~\cite{su2024protrek} extended this to a tri-modal setting incorporating sequences, structures, and textual annotations, highlighting structural complementarity. Inspired by CLIP~\cite{radford2021clip}, methods including ProtCLIP~\cite{zhou2025protclip} and ProteinCLIP~\cite{wu2024proteinclip} further paired large-scale protein language models with curated text to yield semantically coherent, function-aware embeddings. Text-informed frameworks have also proven effective in specialized tasks: MMSite~\cite{ouyang2024mmsite} for active-site identification, BioT5~\cite{pei2023biot5} for chemical–language pretraining, and CoSEF-DBP~\cite{zhang2025cosef} and ProteinDT~\cite{liu2025text} for DNA-binding protein identification and protein design. Collectively, these studies show that textual knowledge provides crucial signals for function annotation, structure–function reasoning, and generative protein engineering.
In this work, we follow this paradigm and introduce a textual quality-control mechanism to address reliability challenges in language-based supervision, thereby improving performance and generalization in enzyme–reaction retrieval.

\begin{figure*}[t]
    \centering
    \includegraphics[width=\linewidth]{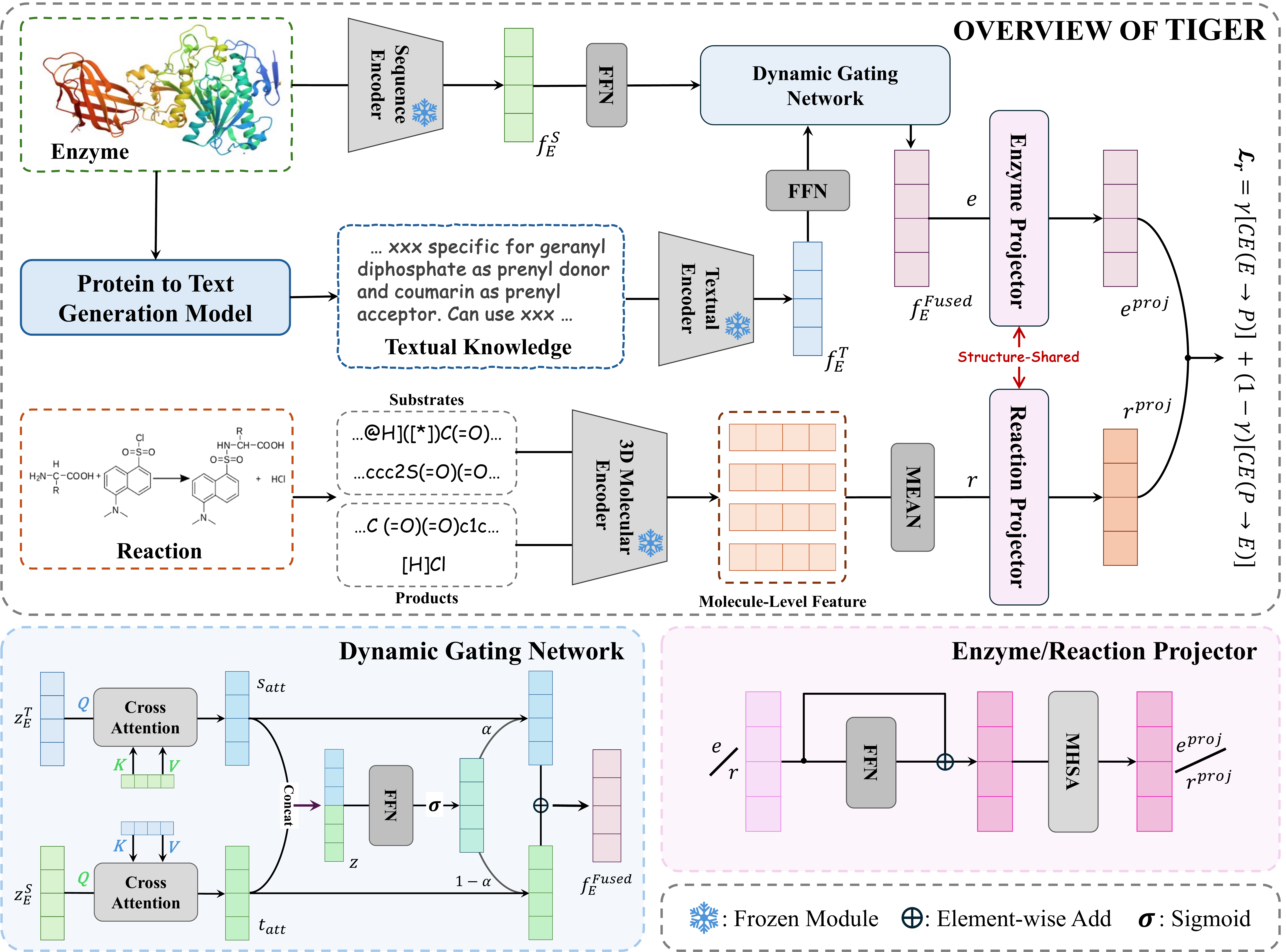}
    \caption{Overview of the proposed TIGER framework. Enzyme sequences and generated textual Knowledge are adaptively fused by a Dynamic Gating Network, while reactions are represented through a 3D molecular encoder. The two modalities are projected into a shared embedding space via Structure-Shared Feature Projectors and jointly optimized with bidirectional contrastive learning for generalized enzyme–reaction retrieval.}
    \label{fig:Overview}
\end{figure*}

\section{Our Approach}
TIGER is fundamentally designed under a contrastive learning paradigm. As shown in Figure~\ref{fig:Overview}, the framework mainly consists of two branches: multimodal enzyme representation learning and reaction representation learning, which are jointly optimized through bidirectional contrastive learning to capture their underlying semantic correspondence. On the enzyme side, sequence embeddings from a pre-trained protein language model are fused with textual semantics via a Dynamic Gating Network to balance complementary information and suppress noise. On the reaction side, 3D molecular encoders extract structural representations of substrates and products, which are aggregated into reaction-level embeddings. Both branches are mapped into a unified embedding space through Structure-Shared Feature Projectors, ensuring symmetric alignment and robust generalization for retrieval.

\subsection{Enzyme Representation Learning}
For any enzyme $e \in \mathcal{E}$, we denote its amino acid sequence as $s_e$. Based on $s_e$, we obtain an automatically generated textual knowledge $t_e$ through a protein-to-text generation model. To construct a robust enzyme representation, we fuse the sequence embedding and textual embedding using a Dynamic Gating Network, which adaptively balances complementary information while suppressing noise from generated text.

\subsubsection{Multimodal Feature Extracting}
For each enzyme $e \in \mathcal{E}$, we derive modality-specific representations from both sequence and textual views. The amino acid sequence $s_e$ is encoded by the pretrained protein language model ESM2~\cite{lin2023esm2}, which effectively captures contextualized dependencies:
\[
f^S_E = \psi_{\text{seq}}(s_e),
\]
while the automatically generated textual description $t_e$ is embedded using PubMedBERT~\cite{gu2021domain}, a domain-specific language model trained on large-scale biomedical literature:
\[
f^T_E = \psi_{\text{text}}(t_e).
\]
To obtain task-adaptive and dimensionally consistent representations, both embeddings are further transformed through modality-specific feed-forward networks:
\[
z^S_e = \mathrm{FFN}_S(f^S_E), \quad 
z^T_e = \mathrm{FFN}_T(f^T_E).
\]
In this way, we obtain two complementary features $z^S_e$ and $z^T_e$, where the former emphasizes structural and sequential context, and the latter conveys functional and semantic information. These features jointly form the foundation for unified enzyme embeddings.

\begin{figure*}[t]
  \centering
  \begin{subfigure}[t]{0.45\linewidth}
    \centering
    \includegraphics[width=\linewidth]{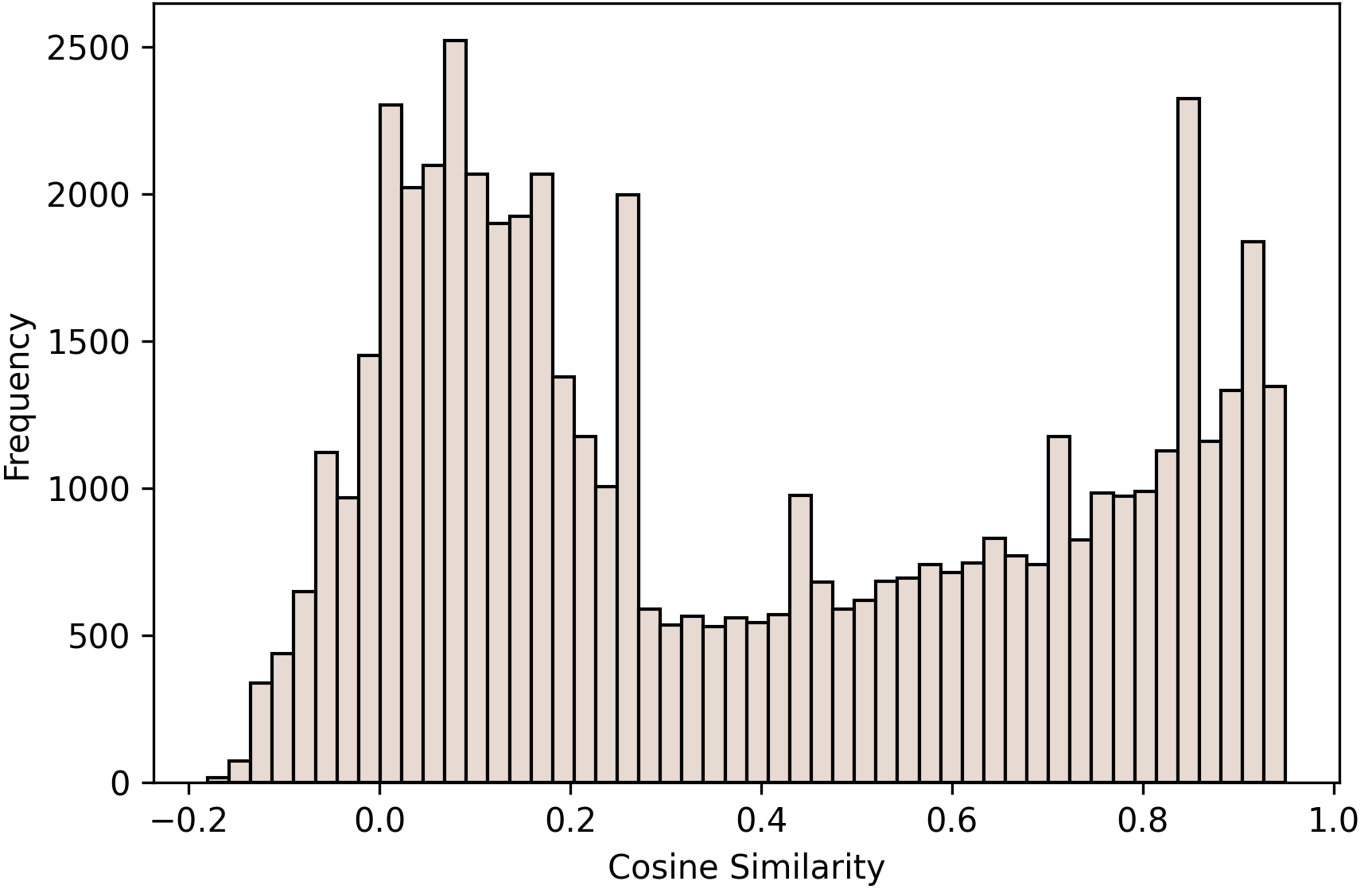}
    \caption{ESM2Text model}
    \label{fig:cos_text_esm}
  \end{subfigure}
  \hspace{0.05\linewidth}
  \begin{subfigure}[t]{0.45\linewidth}
    \centering
    \includegraphics[width=\linewidth]{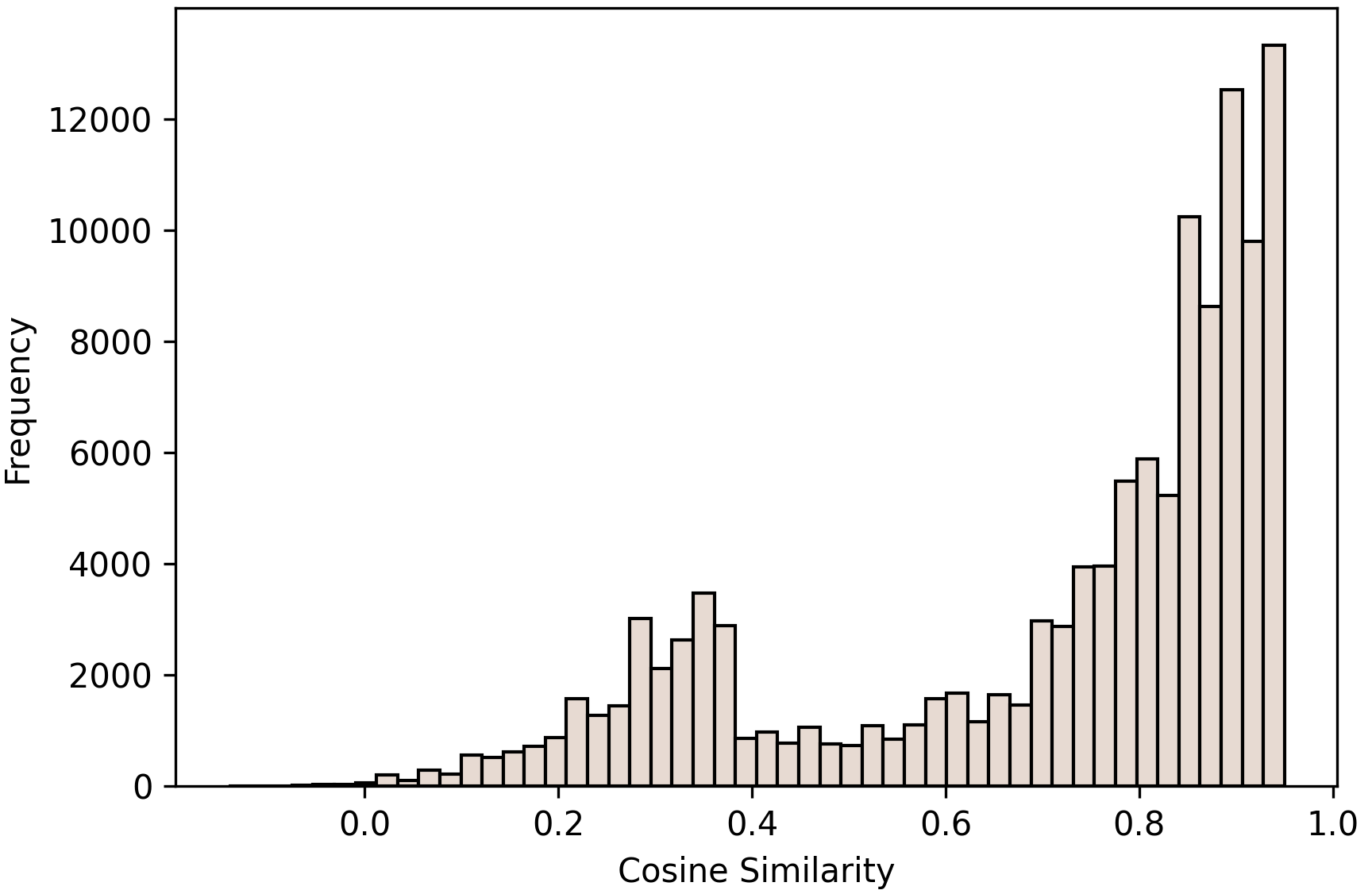}
    \caption{ProtT3 model}
    \label{fig:cos_text_prott3}
  \end{subfigure}

  \caption{Cosine similarities between AI-generated textual knowledge and human-reviewed annotations.}
  \label{fig:cos_text}
\end{figure*}
\subsubsection{Dynamic Gating Network}  
We computed the distribution of cosine similarities between AI-generated textual knowledge (produced by ESM2Text and ProtT3) and human-reviewed SwissProt~\cite{bairoch2000swiss} descriptions.
As illustrated in Figure~\ref{fig:cos_text}, we place particular emphasis on samples with similarity scores below $0.95$.
We observe that approximately one-third of ESM2Text samples and nearly two-thirds of ProtT3 samples fall below $0.95$, indicating that while the majority of generated descriptions align well with curated references, a substantial fraction exhibits noticeable semantic divergence.

Such discrepancies introduce noise into cross-modal alignment and may undermine downstream performance, thereby highlighting the importance of text quality control. To this end, we propose the \textit{Dynamic Gating Network}, a reliability-aware integration mechanism that adaptively modulates the contribution of textual features according to their estimated quality, ultimately improving the robustness of multimodal enzyme representations.

Building on this motivation, the Dynamic Gating Network operates on the features $z^S_e$ and $z^T_e$, progressively integrating them through cross attention and adaptive gating. We first employ bidirectional multi-head attention to enable semantic refinement across modalities:  
\[
s_{att} = \mathrm{MHA}(z^S_e, z^T_e, z^T_e),~
t_{att} = \mathrm{MHA}(z^T_e, z^S_e, z^S_e).
\]  
The attended features are then combined via a gating mechanism that estimates their relative reliability. Specifically, a joint representation is constructed as  
\[
z = [s_{att} \,\|\, t_{att}],
\]  
from which a gating coefficient is derived:  
\[
\boldsymbol{\alpha} = \sigma(W_g z),
\]  
where $\sigma$ denotes the sigmoid function. The gated fusion is computed as  
\[
f_{gated} = \boldsymbol{\alpha} \odot s_{att} + (1 - \boldsymbol{\alpha}) \odot t_{att}.
\]  
Finally, to stabilize the integration, we concatenate $f_{gated}$ with the aggregated signal $(s_{att} + t_{att})$ and apply a feed-forward transformation:  
\[
f^{Fused}_E = \mathrm{FFN}_{fuse}\Big([f_{gated} \,\|\, (s_{att} + t_{att})]\Big).
\]  
The resulting representation $f^{Fused}_E \in \mathbb{R}^{d}$ serves as the unified enzyme representation, providing a robust and reliability-aware basis for contrastive learning against reaction representations.

\subsection{Reaction Representation Learning.}  
For each biochemical reaction $r \in \mathcal{R}$, we follow prior studies that have demonstrated the effectiveness of molecular pre-trained models in capturing reaction semantics, and adopt UniMol-3D~\cite{maziarka2020molecule}, one of the most widely used and high-performing molecular encoders. Specifically, the reaction is decomposed into its constituent substrates and products. Each molecule is independently encoded by UniMol-3D, which leverages both graph-level and 3D conformational information to generate chemically meaningful representations. This strategy ensures that stereochemical and geometric cues, which are often critical for catalytic processes, are faithfully preserved. The resulting molecular embeddings are then aggregated to form the reaction-level representation. Concretely, we compute the reaction embedding by averaging over all encoded substrates and products:
\[
\mathbf{r} = \frac{1}{|\mathcal{S}| + |\mathcal{P}|}
\sum_{x \in \mathcal{S} \cup \mathcal{P}} \mathrm{UniMol}(x),
\]

where $\mathcal{S}$ and $\mathcal{P}$ denote the sets of substrates and products, respectively. This design choice, consistent with the standard practice in recent benchmark works, provides a simple yet robust way to derive reaction features that capture both local molecular structure and global reaction context.

The extracted reaction representation and the enzyme representation introduced are subsequently projected into a shared latent space via the Structure-Shared Feature Projector. This architecture enables cross-modal alignment under contrastive supervision and facilitates bidirectional retrieval between enzymes and reactions.

\subsection{Structure-Shared Feature Projector}
To enable effective enzyme-reaction retrieval, we introduce the Structure-Shared Feature Projector, a dual-branch module that maps heterogeneous inputs into a unified embedding space. Each modality is transformed through a symmetric pipeline of non-linear encoding, residual connections, attention-based contextualization, and final projection:
\[
\phi(\mathbf{x}) = \mathrm{LN}_{proj}\!\left(\mathrm{MHSA}\!\left(\mathrm{FFN}(\mathbf{x}) + \mathbf{W}_{res}\mathbf{x}\right)\right),
\]
where $\mathbf{x} \in \{\mathbf{e}, \mathbf{r}\}$, $\mathrm{LN}_{proj}$ ensures dimensional consistency, and $\mathbf{W}_{res}$ provides residual enhancement. The projected embeddings $\phi(\mathbf{e})$ and $\phi(\mathbf{r})$ lie in a shared space $\mathbb{R}^d$, with a learnable temperature $\tau$ scaling pairwise similarities during contrastive supervision.  

This design enforces semantic proximity between enzymes and reactions, enabling robust bidirectional retrieval.


\subsection{Contrastive Training Objective}
We employ a symmetric contrastive learning objective to align enzyme and reaction representations in a shared latent space. Given a batch of $N$ enzyme–reaction pairs $\{(\mathbf{e}_i, \mathbf{r}_i)\}_{i=1}^{N}$, the similarity between projected embeddings is defined as
\[
s_{ij} = \frac{\mathbf{e}_i^{proj} \cdot \mathbf{r}_j^{proj}}{\tau \|\mathbf{e}_i^{proj}\|\|\mathbf{r}_j^{proj}\|},
\]
where $\tau > 0$ is a learnable temperature parameter. The bidirectional losses are
\[
\mathcal{L}_{e2r} = -\frac{1}{N} \sum_{i=1}^{N} \log \frac{\exp(s_{ii})}{\sum_{j=1}^{N} \exp(s_{ij})},
\]
\[
\mathcal{L}_{r2e} = -\frac{1}{N} \sum_{i=1}^{N} \log \frac{\exp(s_{ii})}{\sum_{j=1}^{N} \exp(s_{ji})},
\]
and the final retrieval loss is
\[
\mathcal{L}_{r} = \gamma \mathcal{L}_{e2r} + (1-\gamma)\mathcal{L}_{r2e},
\]
where $\gamma$ balances two retrieval directions .

\begin{table*}[t]
\centering
\resizebox{\linewidth}{!}{%
\begin{tabular}{c|cc|cc|cc|cc|cc|cc}
\toprule
Split & \multicolumn{4}{c|}{Time-based Split} 
 & \multicolumn{4}{c|}{Enzyme Similarity-based Split} 
 & \multicolumn{4}{c}{Reaction Similarity-based Split} \\
\cmidrule(lr){1-1}\cmidrule(lr){2-5} \cmidrule(lr){6-9} \cmidrule(lr){10-13}
Direction & \multicolumn{2}{c|}{E$\to$R} & \multicolumn{2}{c|}{R$\to$E}
 & \multicolumn{2}{c|}{E$\to$R} & \multicolumn{2}{c|}{R$\to$E}
 & \multicolumn{2}{c|}{E$\to$R} & \multicolumn{2}{c}{R$\to$E} \\
\cmidrule(lr){1-1}\cmidrule(lr){2-5} \cmidrule(lr){6-9} \cmidrule(lr){10-13}
Method & Hit@1 & MRR & Hit@1 & MRR & Hit@1 & MRR & Hit@1 & MRR & Hit@1 & MRR & Hit@1 & MRR \\
\midrule
ReactZyme\textsuperscript{\romannumeral 1} & 0.291 & 0.410 & 0.168 & 0.140 & 0.727 & 0.811 & 0.409 & 0.293 & 0.091 & 0.201 & 0.135 & 0.134 \\
ReactZyme\textsuperscript{\romannumeral 2} & 0.325 & 0.218 & 0.218 & 0.179 & 0.599 & 0.728 & 0.362 & 0.259 & 0.109 & 0.199 & 0.093 & 0.096 \\
ReactZyme\textsuperscript{\romannumeral 3}& 0.092 & 0.159 & 0.056 & 0.054 & 0.600 & 0.723 & 0.348 & 0.256 & 0.094 & 0.194 & 0.114 & 0.104 \\
Fingerprint & 0.236 & 0.298 & 0.144 & 0.117 & 0.579 & 0.639 & 0.255 & 0.204 & 0.094 & 0.194 & 0.114 & 0.104 \\
GNN\textsuperscript{\romannumeral 1} & 0.359 & 0.495 & 0.205 & 0.163 & 0.711 & 0.802 & 0.393 & 0.284 & 0.110 & 0.201 & 0.124 & 0.113 \\
GNN\textsuperscript{\romannumeral 3}& 0.251 & 0.345 & 0.133 & 0.112 & 0.633 & 0.746 & 0.366 & 0.263 & 0.096 & 0.197 & 0.092 & 0.105 \\
Bi-RNN\textsuperscript{\romannumeral 1} & 0.354 & 0.494 & 0.254 & 0.211 & 0.811 & 0.875 & 0.509 & 0.387 & 0.109 & 0.197 & 0.124 & 0.121 \\
Bi-RNN\textsuperscript{\romannumeral 2} & 0.391 & 0.530 & 0.265 & 0.227 & 0.815 & 0.886 & 0.589 & 0.456 & 0.118 & 0.240 & 0.171 & 0.170 \\
CLIPZyme\textsuperscript{\romannumeral 1} & 0.263 & 0.394 & 0.133 & 0.131 & 0.755 & 0.855 & 0.357 & 0.283 & 0.131 & 0.194 & 0.130 & 0.125 \\
CLIPZyme\textsuperscript{\romannumeral 2} & 0.304 & 0.436 & 0.176 & 0.168 & 0.549 & 0.697 & 0.334 & 0.204 & 0.124 & 0.220 & 0.146 & 0.152 \\
\midrule
TIGER (ESM2Text) & \underline{0.581} & \textbf{0.690} & \textbf{0.454} & \underline{0.366} & \textbf{0.931} & \textbf{0.956} & \textbf{0.792}& \textbf{0.592} & \textbf{0.416} & \textbf{0.518} & \textbf{0.430} & \underline{0.319} \\
TIGER (ProtT3) & \textbf{0.583} & \underline{0.683} & \textbf{0.454} & \textbf{0.372} & \underline{0.908} & \underline{0.940} & \underline{0.784} & \underline{0.579} & \underline{0.386} & \underline{0.472} & \underline{0.428} & \textbf{0.337} \\
\bottomrule
\end{tabular}%
}
\caption{Performance comparison across three splits on ReactZyme. The encoders used by the baselines: \textsuperscript{\romannumeral 1}UniMol-3D + ESM, \textsuperscript{\romannumeral 2}MAT-2D + ESM, \textsuperscript{\romannumeral 3}UniMol-3D + SaProt. The best-performing result for each metric is highlighted in \textbf{bold}, while the second-best result is indicated by \underline{underline}.}
\label{tab:three_splits}
\end{table*}

\begin{figure*}[t]
    \centering
    \begin{minipage}{0.32\linewidth}
        \centering
        \includegraphics[width=\linewidth]{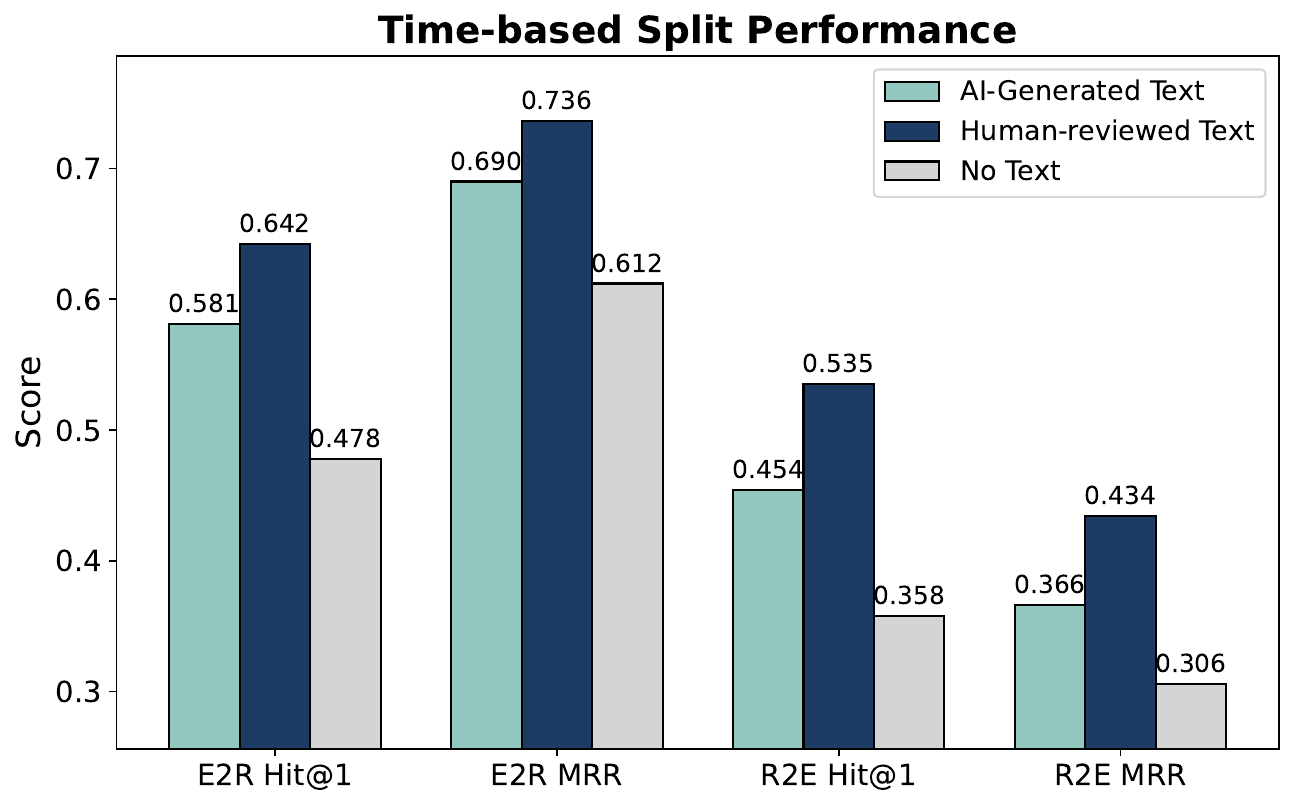}
    \end{minipage}
    \hfill
    \begin{minipage}{0.32\linewidth}
        \centering
        \includegraphics[width=\linewidth]{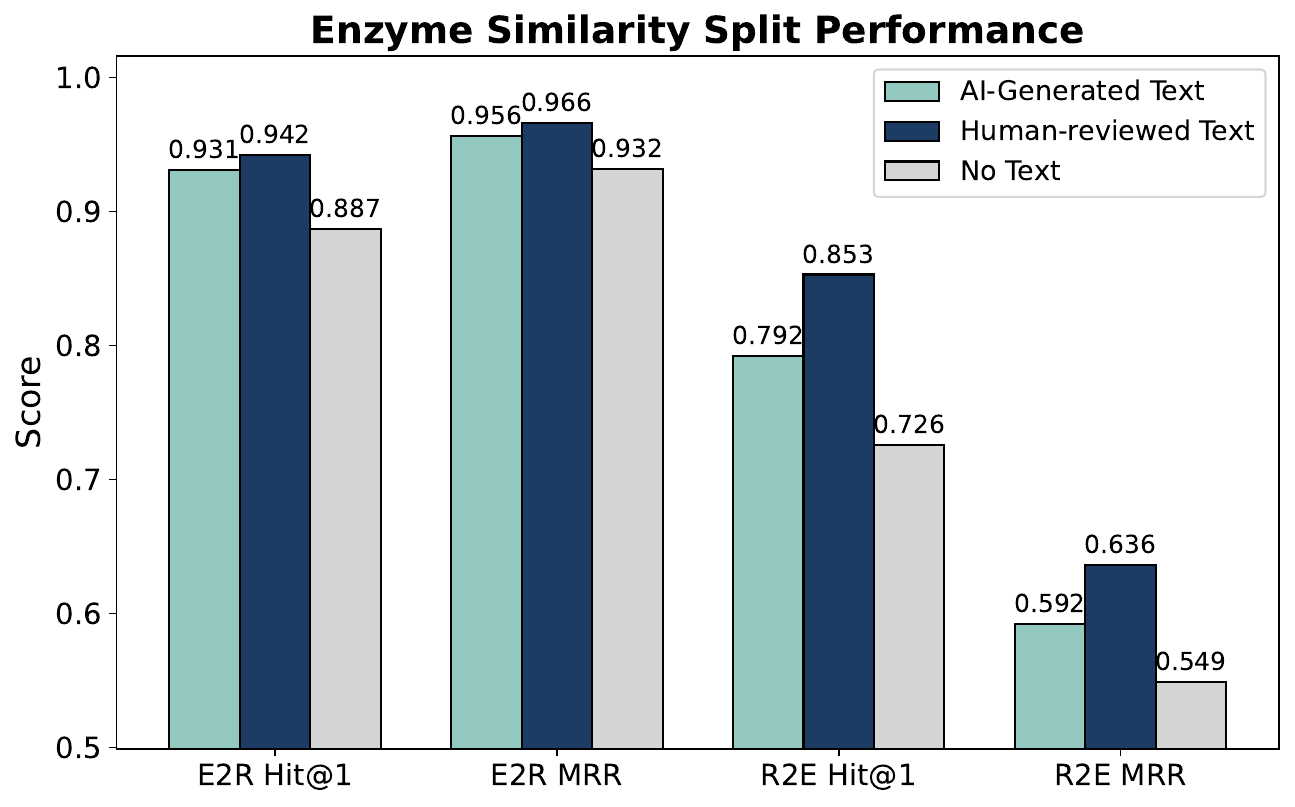}
    \end{minipage}
    \hfill
    \begin{minipage}{0.32\linewidth}
        \centering
        \includegraphics[width=\linewidth]{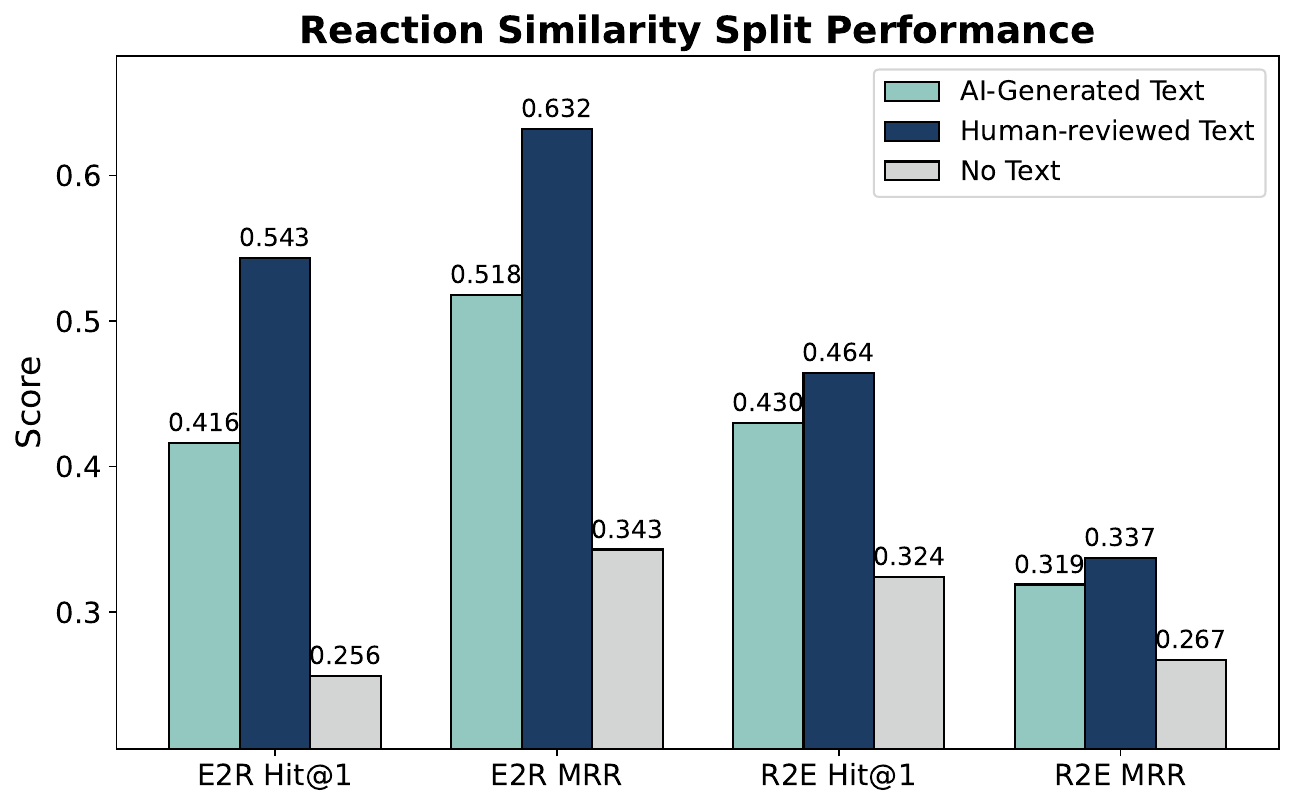}
    \end{minipage}
    \caption{
        Performance comparison across three evaluation splits under different textual settings.
    }
    \label{fig:three_splits}
\end{figure*}

\section{Experiments and Analysis}

\subsection{Dataset: ReactZyme}
\textbf{ReactZyme~\cite{hua2024reactzyme}} is the latest and most comprehensive benchmark for enzyme–reaction retrieval, constructed from curated SwissProt and Rhea resources. It contains over 178K enzyme–reaction associations, spanning more than 178K unique enzymes and 7.7K distinct reactions, thereby providing a functionally grounded alternative to traditional EC- or ontology-based annotations. To rigorously assess generalization, ReactZyme defines three complementary evaluation splits: \textbf{time-based}, where training pairs precede a temporal cutoff while later pairs are reserved for testing; \textbf{enzyme similarity–based}, where test enzymes are sequence-dissimilar to those in training; and \textbf{reaction similarity–based}, where test reactions are entirely absent from training. These settings form a progressive hierarchy of difficulty, with the reaction similarity split posing the greatest challenge as it requires extrapolation to unseen chemical transformations.
  
\subsection{Results Analysis}  
Our evaluation on the ReactZyme dataset examines how TIGER improves retrieval performance and generalization compared to existing baselines. We further analyze how textual knowledge, together with mechanisms such as the Dynamic Gating Network and the loss setting, contributes to the overall effectiveness of the framework. The following results provide quantitative evidence for these improvements.\footnote{Implementation details can be found in the Appendix(\ref{sec:ID}).}
\subsubsection{Performance Analysis of TIGER}
Table~\ref{tab:three_splits} presents a comprehensive comparison between TIGER and representative baselines across three evaluation splits. 
For clarity, the superscripts denote the encoder configurations used by the baselines under the ReactZyme protocol: 
\textsuperscript{\romannumeral 1}UniMol-3D for reactions combined with ESM for enzymes, 
\textsuperscript{\romannumeral 2}MAT-2D with ESM, 
and \textsuperscript{\romannumeral 3}UniMol-3D with SaProt. 
These combinations are widely adopted in molecular representation learning and thus serve as strong reference settings for test.

From the quantitative results, three salient observations can be drawn. 
First, in terms of absolute accuracy, \textbf{TIGER consistently surpasses all baseline methods across all evaluation splits.}
Under the time-based split, TIGER improves Hit@1 from $0.391$ to $0.583$ in enzyme-to-reaction retrieval and from $0.265$ to $0.454$ in the reverse direction compared to the strongest baseline. This advantage persists across other evaluation regimes: on the enzyme similarity-based split, TIGER attains up to $0.931/0.792$ Hit@1 (E$\rightarrow$R / R$\rightarrow$E), and on the most challenging reaction similarity-based split, it achieves $0.416/0.430$ Hit@1. In addition to top-$1$ accuracy, TIGER consistently yields higher MRR scores across all splits and directions, indicating superior overall ranking quality under diverse and heterogeneous evaluation conditions.

Second, \textbf{TIGER exhibits strong bidirectional consistency across all evaluation splits.}
In contrast to many baseline methods that suffer from pronounced directional asymmetry, TIGER maintains balanced performance between enzyme-to-reaction and reaction-to-enzyme retrieval. For example, on the enzyme similarity-based split, the gap between E$\rightarrow$R and R$\rightarrow$E Hit@1 is limited to $0.139$ ($0.931$ vs. $0.792$), while both directions achieve high MRR scores ($0.956/0.592$). Similar patterns are observed under the time-based and reaction similarity-based splits, where TIGER consistently preserves comparable accuracy and ranking quality across directions. These results indicate that TIGER constructs a more coherent and symmetric shared embedding space for enzymes and reactions, effectively alleviating the intrinsic directional bias present in prior approaches.


Third, \textbf{TIGER demonstrates strong robustness across heterogeneous evaluation conditions.}
As evaluation difficulty increases from enzyme similarity-based to time-based and ultimately to reaction similarity-based splits, baseline methods exhibit severe performance collapse, with E$\rightarrow$R Hit@1 typically dropping from $>0.6$ to below $0.15$ in the most challenging setting. In contrast, TIGER degrades much more gracefully, maintaining Hit@1 above $0.41$ in both E$\rightarrow$R and R$\rightarrow$E retrieval on the reaction similarity-based split (0.416 / 0.430). This corresponds to nearly a fourfold improvement over the strongest baseline and underscores TIGER’s superior generalization to temporally unseen enzymes and structurally dissimilar reactions.


The performance gains of TIGER are more clearly illustrated in Figure~\ref{fig:Spider}. 
For brevity, Table~\ref{tab:three_splits} reports only Hit@1 and MRR, while a more comprehensive evaluation, including Hit@K at multiple cutoffs, Precision@K, and Mean Rank, is deferred to the Appendix~\ref{sec:cr}, where we provide a detailed analysis across diverse evaluation dimensions.

\begin{table*}[t]
\centering
\resizebox{\linewidth}{!}{%
\begin{tabular}{c|c|cc|cc|cc|cc|cc|cc}
\toprule
\multicolumn{2}{c|}{\multirow{2}{*}{Setting}}
& \multicolumn{4}{c|}{Time-based Split}
& \multicolumn{4}{c|}{Enzyme Similarity-based Split}
& \multicolumn{4}{c}{Reaction Similarity-based Split} \\
\cmidrule(lr){3-6} \cmidrule(lr){7-10} \cmidrule(lr){11-14}
\multicolumn{2}{c|}{}
& \multicolumn{2}{c|}{E2R} & \multicolumn{2}{c|}{R2E}
& \multicolumn{2}{c|}{E2R} & \multicolumn{2}{c|}{R2E}
& \multicolumn{2}{c|}{E2R} & \multicolumn{2}{c}{R2E} \\
\cmidrule(lr){1-2} \cmidrule(lr){3-14}
Text Source & DGN
& Hit@1 & MRR & Hit@1 & MRR
& Hit@1 & MRR & Hit@1 & MRR
& Hit@1 & MRR & Hit@1 & MRR \\
\midrule
ESM2Text & w/  
& \textbf{0.581} & \textbf{0.690} & \textbf{0.454} & \textbf{0.366}
& \textbf{0.931} & \textbf{0.956} & \textbf{0.792} & \textbf{0.592}
& \textbf{0.416} & \textbf{0.518} & \textbf{0.430} & \textbf{0.319} \\
ESM2Text & w/o 
& 0.531 & 0.646 & 0.395 & 0.319
& 0.912 & 0.945 & 0.760 & 0.566
& 0.391 & 0.482 & 0.389 & 0.296 \\
\midrule
ProtT3 & w/  
& \textbf{0.583} & \textbf{0.683} & \textbf{0.454} & \textbf{0.372}
& \textbf{0.908} & \textbf{0.940} & \textbf{0.784} & \textbf{0.579}
& \textbf{0.386} & \textbf{0.472} & \textbf{0.428} & \textbf{0.337} \\
ProtT3 & w/o 
& 0.529 & 0.638 & 0.366 & 0.325
& 0.888 & 0.929 & 0.703 & 0.533
& 0.284 & 0.412 & 0.306 & 0.263 \\
\midrule
SwissProt & w/  
& \textbf{0.642} & \textbf{0.736} & \textbf{0.535} & \textbf{0.434}
& \textbf{0.942} & \textbf{0.966} & \textbf{0.853} & \textbf{0.636}
& \textbf{0.543} & \textbf{0.632} & \textbf{0.464} & 0.337 \\
SwissProt & w/o 
& 0.572 & 0.692 & 0.456 & 0.376
& 0.928 & 0.958 & 0.802 & 0.601
& 0.403 & 0.504 & 0.427 & \textbf{0.346} \\
\bottomrule
\end{tabular}}
\caption{Performance comparison across three evaluation splits between AI-generated text (ESM2Text, ProtT3) and human-reviewed text (SwissProt), with and without the Dynamic Gating Network (DGN).}
\label{tab:dgn_text_settings}
\end{table*}

\subsubsection{Effect Analysis of Textual Knowledge and Dynamic Gating Network}

\textbf{Effect of Textual Knowledge.} To further investigate the impact of text and its quality, we evaluated three settings: 
\textit{AI-generated text} from ESM2Text, \textit{human-reviewed text} from SwissProt, and \textit{no text}. 
Since SwissProt annotations could be considered as additional resources, the results reported in the main comparison rely on AI-generated descriptions. As shown in Figure~\ref{fig:three_splits}, incorporating text consistently improves retrieval across all splits. 
In the Time-based split (E2R), Hit@1 increases from 0.478 (no text) to 0.581 (AI) and 0.642 (human), while MRR rises from 0.612 $\rightarrow$ 0.690 $\rightarrow$ 0.736. 
Similar trends appear in R2E, with Hit@1 improving from 0.358 to 0.454 and 0.535. 
In the Enzyme Similarity split, both text types bring further gains, with human-reviewed text reaching 0.942 Hit@1 for E2R compared to 0.931 (AI) and 0.887 (no text). 
The Reaction Similarity split shows the largest relative gap: E2R Hit@1 improves from 0.256 to 0.416 and 0.543, and R2E from 0.324 to 0.430 and 0.464. 
These results confirm that textual information provides complementary semantics beyond sequence features, and higher-quality human-curated text delivers consistent additional benefits.



\textbf{Effect of Dynamic Gating Network.} The results in Table~\ref{tab:dgn_text_settings} highlight the significant contribution of the Dynamic Gating Network (DGN). Under the AI-generated text setting, where textual quality can be noisy and unstable (see Figure~\ref{fig:cos_text}), noticeable performance gaps are observed between different text sources, such as ProtT3 and ESM2Text, in the absence of DGN. With DGN enabled, these discrepancies are substantially reduced, and the model achieves more consistent gains across all metrics. For example, in the time-based split, Hit@1 improves from 0.531 to 0.581 (+0.050) and R2E Hit@1 from 0.395 to 0.454 (+0.059); in the enzyme similarity split, E2R Hit@1 rises from 0.912 to 0.931 (+0.019), and R2E MRR from 0.566 to 0.592 (+0.026). A similar trend is observed with SwissProt (human-reviewed) text, where DGN further strengthens performance, improving E2R Hit@1 from 0.572 to 0.642 (+0.070) and R2E MRR from 0.376 to 0.434 (+0.058) in the time-based split, and increasing E2R MRR from 0.504 to 0.632 (+0.128) in the reaction similarity split.
Overall, these results suggest that DGN helps suppress noisy or redundant textual signals and reduces sensitivity to text source quality. This leads to more stable and robust text-informed retrieval across different experimental settings.

\subsubsection{Ablation Study on Structure-Shared Feature Projector}
\label{sec:SSFP}

\begin{table*}[t]
\centering
\resizebox{\linewidth}{!}{%
\begin{tabular}{c|cc|cc|cc|cc|cc|cc}
\toprule
\multicolumn{1}{c|}{\multirow{2}{*}{Method}}
& \multicolumn{4}{c|}{Time-based Split}
& \multicolumn{4}{c|}{Enzyme Similarity-based Split}
& \multicolumn{4}{c}{Reaction Similarity-based Split} \\
\cmidrule(lr){2-5} \cmidrule(lr){6-9} \cmidrule(lr){10-13}
& \multicolumn{2}{c|}{E$\rightarrow$R} & \multicolumn{2}{c|}{R$\rightarrow$E}
& \multicolumn{2}{c|}{E$\rightarrow$R} & \multicolumn{2}{c|}{R$\rightarrow$E}
& \multicolumn{2}{c|}{E$\rightarrow$R} & \multicolumn{2}{c}{R$\rightarrow$E} \\
\cmidrule(lr){2-13}
& Hit@1 & MRR & Hit@1 & MRR
& Hit@1 & MRR & Hit@1 & MRR
& Hit@1 & MRR & Hit@1 & MRR \\
\midrule
No SSFP
& 0.515 & 0.635 & \textbf{0.461} & 0.365
& 0.901 & 0.938 & \textbf{0.814} & 0.606
& 0.374 & 0.489 & 0.386 & 0.273 \\

2-layer MLP
& 0.565 & 0.674 & 0.438 & 0.352
& 0.914 & 0.948 & 0.813 & \textbf{0.607}
& \textbf{0.446} & \textbf{0.543} & 0.383 & 0.289 \\

SSFP (Ours)
& \textbf{0.581} & \textbf{0.690} & 0.454 & \textbf{0.366}
& \textbf{0.931} & \textbf{0.956} & 0.792 & 0.592
& 0.416 & 0.518 & \textbf{0.430} & \textbf{0.319} \\
\bottomrule
\end{tabular}}
\caption{Ablation study on the Structure-Shared Feature Projector (SSFP) across different evaluation splits and retrieval directions.}
\label{tab:ssfp}
\end{table*}



The Structure-Shared Feature Projector (SSFP) is introduced as a functional projection module to map enzyme and reaction representations into a unified feature space. To evaluate its effectiveness, we compare SSFP against two variants: removing the projector entirely (No SSFP) and replacing it with a standard two-layer MLP, with results summarized in Table~\ref{tab:ssfp}.

The results demonstrate that SSFP enhances retrieval performance across most scenarios, though its impact varies depending on the distribution shift. On the time-based split, SSFP significantly improves E$\rightarrow$R retrieval, achieving the highest Hit@1 (0.581) and MRR (0.690). While the R$\rightarrow$E Hit@1 is competitive with the baselines, SSFP still secures the best R$\rightarrow$E MRR (0.366), indicating a better overall ranking of correct candidates.
Furthermore, SSFP shows distinct advantages under specific structural shifts. On the enzyme similarity-based split, SSFP delivers the strongest E$\rightarrow$R performance (Hit@1 of 0.931 and MRR of 0.956), effectively aligning novel enzymes to known reactions. Conversely, on the reaction similarity-based split, SSFP substantially outperforms both ablated settings in the R$\rightarrow$E direction (Hit@1 of 0.430 and MRR of 0.319), whereas the two-layer MLP proves more effective for E$\rightarrow$R. 

Overall, while simpler projection strategies may suffice for specific single-direction retrievals under narrow shifts, the proposed SSFP provides a more robust and balanced cross-modal alignment. It effectively mitigates strong distribution shifts, validating its necessity in the overall architecture.

\subsubsection{Sensitivity Analysis of $\mathcal{L}_r$}

\begin{figure} 
  \centering
  \includegraphics[width=\linewidth]{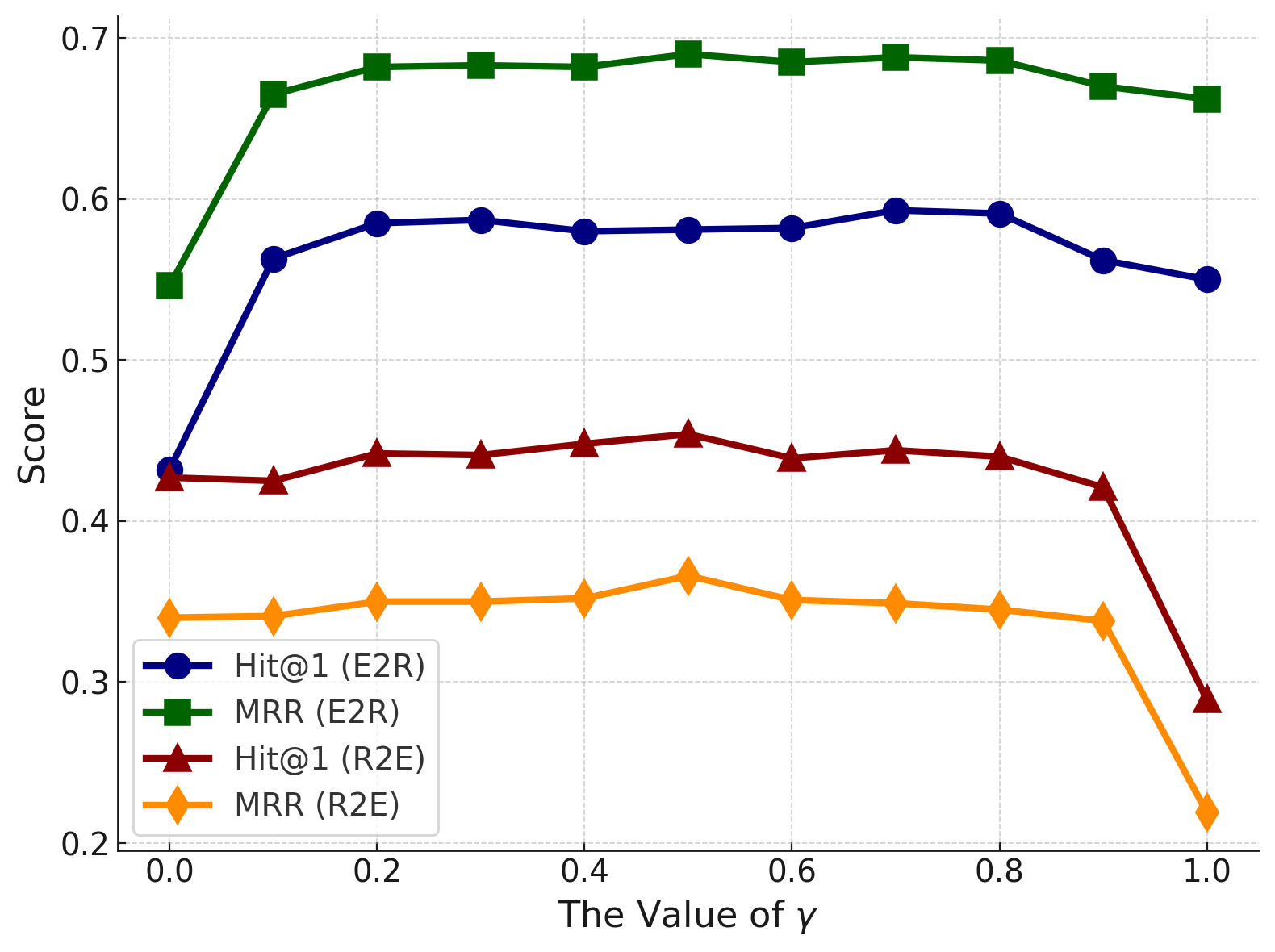}
  \caption{Sensitivity of retrieval performance with respect to the balancing parameter $\gamma$ on the time-based split.}
  \label{fig:gamma_sensitivity}
\end{figure}

We further investigate the influence of the balancing coefficient $\gamma$ on the retrieval objective $\mathcal{L}_r$ by conducting a sensitivity analysis on the time-based split. As depicted in Figure~\ref{fig:gamma_sensitivity}, the performance remains relatively stable when $\gamma$ lies within a moderate range, whereas extreme values cause pronounced degradation. In particular, assigning balanced weights to both retrieval directions ($\gamma \in [0.3,0.7]$) leads to consistently superior results, with $\gamma=0.7$ achieving the highest enzyme-to-reaction Hit@1 score of $0.593$. In contrast, skewed weighting substantially compromises the opposite retrieval direction: for example, $\gamma=1.0$ maximizes the enzyme-to-reaction branch but reduces the reaction-to-enzyme Hit@1 score drastically to $0.289$. These results highlight the necessity of maintaining bidirectional balance in $\mathcal{L}_r$, and a near-symmetric weighting scheme proves preferable for robust retrieval performance. Therefore, we adopt $\gamma=0.5$ as the default setting in all subsequent comparisons, which not only provides a fair balance between the two retrieval directions but also enhances the stability and generalization of the model across different evaluation scenarios.

\section{Conclusion}
In this work, we introduced TIGER, a text-informed generalized enzyme–reaction retrieval framework that addresses the fundamental challenges of directional asymmetry and distributional sensitivity in existing approaches. By leveraging knowledge-rich textual descriptions generated from protein-to-text generation models, TIGER augments sequential representations with functional semantics, while the proposed Dynamic Gating Network ensures reliable integration by suppressing noisy or spurious textual cues. In addition, the Structure-Shared Feature Projector provides a unified embedding space that enhances cross-modal alignment and supports robust bidirectional retrieval. Comprehensive experiments on the ReactZyme benchmark demonstrate that TIGER consistently surpasses strong baselines across time-based, enzyme similarity–based, and reaction similarity–based splits, achieving both improved absolute performance and greater bidirectional consistency. Beyond SOTA results, TIGER highlights the potential of text-informed paradigms for advancing biochemical retrieval tasks. In future work, we plan to extend TIGER towards more fine-grained catalytic annotations, integrate curated biochemical ontologies for richer textual supervision, and explore its applicability in related domains such as metabolic pathway analysis and protein design.

\section{Acknowledgments}
This work was partially supported by Hefei Key Technology Research and Development Project (2024SZD005), and the China National Natural Science Foundation with no. 92567301 and 62132018.

\section{Limitations}
Despite its effectiveness, our framework has several limitations. First, the current modeling of textual descriptions remains relatively coarse-grained, which may fail to fully capture fine-grained catalytic information that is potentially critical for precise retrieval. Second, the quality of textual supervision is only implicitly reflected through downstream performance, and a more explicit mechanism for assessing and modeling text reliability could further improve the robustness of text-informed representations. Addressing these limitations is an important direction for future work.

\bibliography{ref}
\clearpage
\appendix

\section{Ethics Statement}
This study is based exclusively on publicly available biochemical datasets, including ReactZyme and SwissProt, which contain curated information about enzymes and reactions without any personally identifiable or sensitive data. No human or animal subjects are involved. The research is conducted solely for advancing enzyme–reaction retrieval in scientific and educational contexts, and all experiments were designed to ensure transparency, reproducibility, and responsible use of resources.

\section{Use of Large Language Models (LLMs)}
During the preparation of this manuscript, we made limited use of large language models (LLMs) to improve linguistic presentation. 
LLMs were occasionally consulted to refine wording, adjust phrasing, and enhance readability. 
All core components of this work, including the research ideas, methodological design, experimental setup, and analysis, were entirely developed and executed by the authors. 
The involvement of LLMs was strictly confined to language refinement and did not affect the scientific content, technical contributions, or conclusions of the paper.

\section{Implementation details}
\label{sec:ID}
All experiments were implemented in PyTorch and conducted on NVIDIA A40 GPUs. 
We trained the model for 20 epochs using the Adam optimizer with a learning rate of $1\times10^{-3}$. 


\section{Task Formulation}
\label{sec:task}
Let $\mathcal{E} = \{e_1, e_2, \dots, e_N\}$ denote a set of enzyme entities and $\mathcal{R} = \{r_1, r_2, \dots, r_M\}$ denote a set of biochemical reactions. Each enzyme $e_i \in \mathcal{E}$ may be associated with one or more reactions $r_j \in \mathcal{R}$ that it catalyzes or participates in, and vice versa.

We assume access to a partial ground-truth correspondence $\mathcal{A} \subseteq \mathcal{E} \times \mathcal{R}$, where each pair $(e, r) \in \mathcal{A}$ indicates a known biological association. The goal is to recover or rank such associations via representation-based matching in a learned metric space.

Formally, we aim to learn an embedding function
\[
\phi_1 : \mathcal{E} \rightarrow \mathbb{R}^d, \phi_2 : \mathcal{R} \rightarrow \mathbb{R}^d
\]
that maps both enzymes and reactions into a shared latent space $\mathbb{R}^d$ such that semantically associated pairs are close under a similarity metric $s : \mathbb{R}^d \times \mathbb{R}^d \rightarrow \mathbb{R}$. For a matched pair $(e, r) \in \mathcal{A}$, the embedding function should satisfy
\[
s(\phi_1(e), \phi_2(r)) \gg s(\phi_1(e), \phi_2(r')), ~\forall r' \notin \mathcal{A}(e)
\]
and symmetrically,
\[
s(\phi_2(r), \phi_1(e)) \gg s(\phi_2(r), \phi_1(e')), ~\forall e' \notin \mathcal{A}(r).
\]

This problem setting naturally defines a bidirectional retrieval task:
\begin{itemize}
    \item \textbf{Enzyme-to-Reaction (E2R)}: Given an enzyme query $e \in \mathcal{E}$, retrieve the most relevant reaction(s) $r \in \mathcal{R}$ such that $(e, r) \in \mathcal{A}$.
    \item \textbf{Reaction-to-Enzyme (R2E)}: Given a reaction query $r \in \mathcal{R}$, retrieve the most relevant enzyme(s) $e \in \mathcal{E}$ such that $(e, r) \in \mathcal{A}$.
\end{itemize}

This bidirectional matching formulation provides a foundation for downstream applications such as enzyme function prediction, metabolic pathway reconstruction, and biochemical knowledge graph completion. The subsequent sections describe our multimodal representation learning framework and the training procedure used to optimize $\phi_1$ and $\phi_2$ under contrastive supervision.

\section{Baselines under the ReactZyme Protocol}
\label{sec:baselines}

We evaluate our method against representative baselines implemented under the \textbf{ReactZyme} protocol for bidirectional enzyme-reaction retrieval. 
Unless otherwise stated, all methods adopt a dual-encoder architecture with a temperature-scaled cosine similarity and a symmetric InfoNCE-style contrastive objective for both directions (E$\to$R and R$\to$E). 
Superscripts in Table~\ref{tab:three_splits} indicate the specific encoder pairing used by each baseline: 
\textsuperscript{\romannumeral 1} \;UniMol-3D (reaction) + ESM (enzyme), 
\textsuperscript{\romannumeral 2} \;MAT-2D (reaction) + ESM (enzyme), 
\textsuperscript{\romannumeral 3} \;UniMol-3D (reaction) + SaProt (enzyme).

\paragraph{ReactZyme (Base).}
The foundational benchmark that establishes the training/evaluation protocol for enzyme-reaction retrieval.
On the enzyme side, a protein language model (ESM or SaProt per the superscript) encodes amino-acid sequences; on the reaction side, a learned chemical encoder (UniMol-3D or MAT-2D per the superscript) produces reaction embeddings. 
Both embeddings are projected to a shared space via lightweight MLP heads and trained with a bidirectional contrastive loss using in-batch negatives.

\paragraph{Fingerprint.}
A non-neural reaction representation baseline where reactions are encoded by standard chemical fingerprints (RDKit). 
The enzyme encoder follows the ReactZyme setup (ESM or SaProt depending on the variant), and a small MLP is used to map both sides into the shared space. 
This baseline is computationally efficient but typically less expressive for complex reaction semantics.

\paragraph{GNN\textsuperscript{\romannumeral 1, \romannumeral 3}.}
A graph-neural reaction encoder variant in which molecular graphs (and available 3D cues) are processed by a message-passing backbone with a graph-level readout. 
In our runs, the reaction side corresponds to UniMol-3D ($^{1,3}$), paired with ESM (\textsuperscript{\romannumeral 1} ) or SaProt (\textsuperscript{\romannumeral 3} ) on the enzyme side as indicated in the table. 
Projection heads and the contrastive training recipe remain identical to ReactZyme.

\paragraph{Bi\textendash RNN\textsuperscript{\romannumeral 1, \romannumeral 2}.}
As another sequential decoding baseline, we consider a bidirectional recurrent neural network (Bi-RNN) as the decoder. Unlike the simple feed-forward MLP, the Bi-RNN is designed to capture temporal dependencies by processing the encoded representations in both forward and backward directions, thereby modeling contextual interactions across the sequence. In our implementation, the Bi-RNN decoder takes the projected embeddings as input and produces hidden states that are aggregated to form the final retrieval scores. This architecture allows the model to exploit sequential ordering and long-range dependencies, providing a stronger sequential inductive bias compared to the MLP decoder. According to the ReactZyme paper~\cite{hua2024reactzyme}, the Bi-RNN decoder demonstrated the best retrieval performance among the tested alternatives; therefore, in our main experiments we adopt this variant for comparison.

\paragraph{CLIPZyme\textsuperscript{\romannumeral 1, \romannumeral 2} .}
CLIPZyme~\cite{mikhael2024clipzyme} is a CLIP-style dual-encoder framework designed for enzyme-reaction retrieval. 
On the enzyme side, it adopts a protein language model encoder (e.g., ESM), while on the reaction side, it introduces a novel representation by constructing a \emph{pseudo-transition state graph} that connects substrates and products. 
This graph is intended to approximate the intermediate transition state of biochemical reactions, thereby enriching the reaction representation beyond standard molecular encodings.  
Both enzyme and reaction embeddings are projected into a shared latent space, and training is conducted using a contrastive objective similar to the CLIP paradigm.  
In Table~\ref{tab:three_splits}, superscripts \textsuperscript{\romannumeral 1}  and \textsuperscript{\romannumeral 2}  indicate the use of UniMol-3D or MAT-2D as the reaction encoder in place of the pseudo-transition graph for ablation-style comparisons under the ReactZyme protocol.
\subsection{Extra Experiments and Analysis}
To further validate the robustness and generalizability of our framework, we provide an extended set of experiments across three evaluation splits: the \textit{time-based split}, the \textit{enzyme similarity-based split}, and the \textit{reaction similarity-based split}. 
These additional experiments not only complement the main results presented in the previous section but also offer deeper insights into how TIGER performs under different levels of distributional shifts. 
For each split, we analyze both retrieval directions (enzyme-to-reaction and reaction-to-enzyme) with multiple evaluation metrics, including Hit@k, Precision@k, Mean Reciprocal Rank (MRR), and Mean Rank. 
The following subsections summarize the results and provide detailed analyses.

\section{Extra Experiments and Analysis}
\label{sec:cr}
To provide a more comprehensive evaluation of our framework, we report additional experimental results using an extended set of evaluation metrics. 
Beyond the primary metrics presented in the previous section, we further analyze model performance in both retrieval directions (enzyme-to-reaction and reaction-to-enzyme) under the same evaluation splits.
Specifically, we include Hit@k, Precision@k, Mean Reciprocal Rank (MRR), and Mean Rank to offer a more fine-grained assessment of retrieval quality.
The following subsections summarize the results and provide detailed analyses.

\subsection{Evaluation Metrics}

We adopted several widely used retrieval metrics to comprehensively evaluate model performance:  

\paragraph{Hit@K.}  
Hit@K measures whether the ground-truth item appears within the top-$K$ retrieved results. Formally, for a query $q$, let $\mathrm{rank}(q)$ denote the rank position of its ground-truth item. Then  
\[
\mathrm{H@K} = \frac{1}{N} \sum_{i=1}^{N} \mathbb{I}\big[ \mathrm{rank}(q_i) \leq K \big],
\]  
where $N$ is the total number of queries. Hit@K is also equivalent to Top-$K$ accuracy in classification settings.  

\paragraph{Precision@K.}  
Precision@K evaluates how many of the retrieved top-$K$ items are correct. This metric is particularly useful when a query may correspond to multiple valid ground-truth items. It is defined as  
\[
\mathrm{P@K} = \frac{1}{N} \sum_{i=1}^{N} \frac{|\mathrm{Retrieved}(q_i, K) \cap \mathrm{GT}(q_i)|}{K},
\]  
where $\mathrm{Retrieved}(q_i, K)$ is the set of top-$K$ results for query $q_i$, and $\mathrm{GT}(q_i)$ is its ground-truth set.  

\paragraph{Mean Reciprocal Rank (MRR).}  
MRR evaluates ranking quality by rewarding higher scores when the correct item appears earlier in the ranked list:  
\[
\mathrm{MRR} = \frac{1}{N} \sum_{i=1}^{N} \frac{1}{\mathrm{rank}(q_i)}.
\]  

\paragraph{Mean Rank (MR).}  
Mean Rank directly computes the average rank position of the ground-truth items:  
\[
\mathrm{MR} = \frac{1}{N} \sum_{i=1}^{N} \mathrm{rank}(q_i).
\]  
Lower values indicate better performance, as the correct items tend to appear earlier in the ranking.  

\subsection{Analysis of Retrieval Performance on Time-based Splits}
\paragraph{Enzyme-to-Reaction Retrieval (Hit@k and MRR).}
From Table~\ref{tab:hit_te1}, we observe that TIGER markedly outperforms all baseline methods across all cutoff thresholds and in terms of MRR. Specifically, TIGER achieves a Hit@1 of \textbf{0.5810}, representing a relative improvement of nearly 48\% over the strongest baseline (Bi-RNN\textsuperscript{\romannumeral 2} , 0.3911). Similar improvements persist at higher cutoff thresholds: for example, at Hit@20, TIGER reaches \textbf{0.9164}, which significantly exceeds the next-best baseline (Bi-RNN\textsuperscript{\romannumeral 2} , 0.8559). The consistent margins across H@k levels highlight that TIGER is not only more accurate at top-1 retrieval but also ensures stable ranking quality deeper into the candidate list. Moreover, the MRR of \textbf{0.6902} represents a substantial leap over the baseline range (0.2788-0.5303), further validating TIGER’s effectiveness in optimizing rank-sensitive metrics.

\begin{table*}[t]
\centering
\begin{tabular}{l|ccccccc|c}
\toprule
Method & H@1 & H@2 & H@3 & H@4 & H@5 & H@10 & H@20 & MRR \\
\midrule
ReactZyme\textsuperscript{\romannumeral 1}  & 0.2905 & 0.4007 & 0.4563 & 0.4984 & 0.5365 & 0.6586 & 0.7639 & 0.4104 \\
ReactZyme\textsuperscript{\romannumeral 2}  & 0.3246 & 0.4526 & 0.5255 & 0.5700 & 0.6044 & 0.7079 & 0.7972 & 0.4549 \\
ReactZyme\textsuperscript{\romannumeral 3}  & 0.0916 & 0.1328 & 0.1650 & 0.1908 & 0.2134 & 0.2923 & 0.3882 & 0.2788 \\
Fingerprint   & 0.2357 & 0.3470 & 0.3968 & 0.4215 & 0.4684 & 0.5439 & 0.7040 & 0.2984  \\
GNN\textsuperscript{\romannumeral 1}        & 0.3588 & 0.5158 & 0.5919 & 0.6044 & 0.6545 & 0.7815 & 0.8126 & 0.4952   \\
GNN\textsuperscript{\romannumeral 3}        & 0.2508 & 0.3528 & 0.3995 & 0.4016 & 0.4075 & 0.5448 & 0.6421 & 0.3453\\
Bi-RNN\textsuperscript{\romannumeral 1}     & 0.3543 & 0.5112 & 0.5820 & 0.6250 & 0.6563 & 0.7480 & 0.8259 & 0.4946\\
Bi-RNN\textsuperscript{\romannumeral 2}     & 0.3911 & 0.5542 & 0.6170 & 0.6555 & 0.6875 & 0.7847 & 0.8559 & 0.5303  \\
CLIPZyme\textsuperscript{\romannumeral 1}   & 0.2631 & 0.3670 & 0.4189 & 0.4447 & 0.4534 & 0.6444 & 0.7516 & 0.3940\\
CLIPZyme\textsuperscript{\romannumeral 2}   & 0.3041 & 0.4346 & 0.4991 & 0.5610 & 0.5993 & 0.6943 & 0.7840 & 0.4355\\
\midrule
TIGER (Ours)  & \textbf{0.5810} & \textbf{0.7187} & \textbf{0.7678} & \textbf{0.7972} & \textbf{0.8190} & \textbf{0.8740} & \textbf{0.9164} & \textbf{0.6902} \\
\bottomrule
\end{tabular}
\caption{Enzyme to Reaction Retrieval Performance (H@k and MRR) on Time-based Split.}
\label{tab:hit_te1}
\end{table*}
\paragraph{Enzyme-to-Reaction Retrieval (Precision@k and Mean Rank).}
As shown in Table~\ref{tab:prec_te2}, TIGER sustains its advantage when evaluated with precision-oriented metrics. At P@1, TIGER again attains \textbf{0.5810}, clearly outperforming Bi-RNN\textsuperscript{\romannumeral 2}  (0.3911). While precision values naturally decay with larger k, TIGER consistently dominates baselines across all cutoffs. More importantly, TIGER achieves a \textbf{mean rank of 13.33}, which is dramatically lower (better) than those of existing methods, where even the strongest baselines remain above 30-40. This indicates that correct reactions for enzymes are not only placed earlier but are much more concentrated toward the top of the retrieval list under TIGER’s ranking.
\begin{table*}[t]
\centering
\begin{tabular}{l|ccccccc|c}
\toprule
Method & P@1 & P@2 & P@3 & P@4 & P@5 & P@10 & P@20 & Mean Rank \\
\midrule
ReactZyme\textsuperscript{\romannumeral 1}  & 0.2905 & 0.2004 & 0.1522 & 0.1247 & 0.1074 & 0.0659 & 0.0382 & 46.0553 \\
ReactZyme\textsuperscript{\romannumeral 2}  & 0.3246 & 0.2263 & 0.1752 & 0.1425 & 0.1209 & 0.0708 & 0.0399 & 40.4756 \\
ReactZyme\textsuperscript{\romannumeral 3}  & 0.0916 & 0.0664 & 0.0550 & 0.0477 & 0.0426 & 0.0292 & 0.0194 & 168.8244 \\
Fingerprint   & 0.2357  & 0.1736 & 0.1323 & 0.1054 & 0.0937 & 0.0544 & 0.0352 & 89.5675  \\
GNN\textsuperscript{\romannumeral 1}        & 0.3588 & 0.2579 & 0.1973 & 0.1511 & 0.1309 & 0.0781 & 0.0406 & 32.7443 \\
GNN\textsuperscript{\romannumeral 3}        & 0.2508 & 0.1764 & 0.1331 & 0.1004 & 0.0815 & 0.0546 & 0.0321 & 59.8345 \\
Bi-RNN\textsuperscript{\romannumeral 1}     & 0.3543 & 0.2556 & 0.1940 & 0.1563 & 0.1313 & 0.0748 & 0.0413 & 34.6103 \\
Bi-RNN\textsuperscript{\romannumeral 2}     & 0.3911 & 0.2771 & 0.2057 & 0.1639 & 0.1375 & 0.0785 & 0.0428 & 35.2791 \\
CLIPZyme\textsuperscript{\romannumeral 1}   & 0.2631 & 0.1835 & 0.1401 & 0.1112 & 0.0907 & 0.0645 & 0.0376 & 45.3637 \\
CLIPZyme\textsuperscript{\romannumeral 2}   & 0.3041 & 0.2173 & 0.1658 & 0.1399 & 0.1201 & 0.0695 & 0.0392 & 42.3645 \\
\midrule
TIGER (Ours)  & \textbf{0.5810} & \textbf{0.3593} & \textbf{0.2559} & \textbf{0.1993} & \textbf{0.1638} & \textbf{0.0874} & \textbf{0.0458} & \textbf{13.3309} \\
\bottomrule
\end{tabular}
\caption{Enzyme to Reaction Retrieval Performance (P@k and Mean Rank) on Time-based Split.}
\label{tab:prec_te2}
\end{table*}
\paragraph{Reaction-to-Enzyme Retrieval (Hit@k and MRR).}
Table~\ref{tab:hit_tr1} demonstrates similar trends in the reverse retrieval direction. TIGER surpasses all baselines substantially, with a Hit@1 of \textbf{0.4536}, compared to 0.2650 for Bi-RNN\textsuperscript{\romannumeral 2} , the closest competitor. The improvements remain consistent as k increases: TIGER achieves \textbf{0.6708} at Hit@5 and \textbf{0.8477} at Hit@20, surpassing the best baselines by wide margins. Although absolute values are slightly lower than in the enzyme-to-reaction setting (reflecting the greater difficulty of this direction), TIGER still provides strong improvements in MRR (\textbf{0.3658} versus 0.2267), highlighting its robust generalization across both retrieval tasks.
\begin{table*}[t]
\centering
\begin{tabular}{l|ccccccc|c}
\toprule
Method & H@1 & H@2 & H@3 & H@4 & H@5 & H@10 & H@20 & MRR \\
\midrule
ReactZyme\textsuperscript{\romannumeral 1}  & 0.1678 & 0.2240 & 0.2631 & 0.2938 & 0.3155 & 0.3960 & 0.5011 & 0.1400 \\
ReactZyme\textsuperscript{\romannumeral 2}  & 0.2175 & 0.2733 & 0.3144 & 0.3493 & 0.3815 & 0.4924 & 0.6033 & 0.1789 \\
ReactZyme\textsuperscript{\romannumeral 3}  & 0.0558 & 0.0721 & 0.0815 & 0.0883 & 0.0979 & 0.1359 & 0.1918 & 0.0538 \\
Fingerprint   & 0.1435 & 0.2017 & 0.2345 & 0.2656 & 0.2980 & 0.3547 & 0.4582 & 0.1166  \\
GNN\textsuperscript{\romannumeral 1}        & 0.2045 & 0.2835 & 0.3398 & 0.3722 & 0.3792 & 0.4475 & 0.5168 & 0.1628   \\
GNN\textsuperscript{\romannumeral 3}        & 0.1331 & 0.1750 & 0.1886 & 0.1979 & 0.2044 & 0.3365 & 0.4119 & 0.1122\\
Bi-RNN\textsuperscript{\romannumeral 1}     & 0.2540 & 0.3261 & 0.3747 & 0.4024 & 0.4324 & 0.5330 & 0.6481 & 0.2113\\
Bi-RNN\textsuperscript{\romannumeral 2}     & 0.2650 & 0.3470 & 0.3994 & 0.4355 & 0.4704 & 0.5854 & 0.6940 & 0.2267  \\
CLIPZyme\textsuperscript{\romannumeral 1}   & 0.1331 & 0.2034 & 0.2451 & 0.2822 & 0.2993 & 0.3554 & 0.4567 & 0.1313\\
CLIPZyme\textsuperscript{\romannumeral 2}   & 0.1757 & 0.2445 & 0.3062 & 0.3075 & 0.3447 & 0.4555 & 0.5343 & 0.1678\\
\midrule
TIGER (Ours)  & \textbf{0.4536} & \textbf{0.5474} & \textbf{0.6025} & \textbf{0.6427} & \textbf{0.6708} & \textbf{0.7676} & \textbf{0.8477} & \textbf{0.3658} \\
\bottomrule
\end{tabular}
\caption{Reaction to Enzyme Retrieval Performance (H@k and MRR) on Time-based Split.}
\label{tab:hit_tr1}
\end{table*}
\paragraph{Reaction-to-Enzyme Retrieval (Precision@k and Mean Rank).}
Finally, Table~\ref{tab:prec_tr2} shows TIGER’s precision and ranking performance in the reaction-to-enzyme direction. TIGER achieves a P@1 of \textbf{0.4536}, outperforming Bi-RNN\textsuperscript{\romannumeral 2}  (0.2650) by more than 70\%. The performance gap persists across increasing cutoffs, indicating that TIGER can consistently identify relevant enzymes even when more candidates are considered. Crucially, TIGER’s \textbf{mean rank of 45.13} represents a major reduction compared to the 138-700 range of baselines, showing that TIGER drastically shortens the search depth required to find correct enzymes.
\begin{table*}[t]
\centering
\begin{tabular}{l|ccccccc|c}
\toprule
Method & P@1 & P@2 & P@3 & P@4 & P@5 & P@10 & P@20 & Mean Rank \\
\midrule
ReactZyme\textsuperscript{\romannumeral 1}  & 0.1678 & 0.1543 & 0.1443 & 0.1349 & 0.1267 & 0.1002 & 0.0748 & 177.4881 \\
ReactZyme\textsuperscript{\romannumeral 2}  & 0.2175 & 0.2001 & 0.1817 & 0.1688 & 0.1570 & 0.1206 & 0.0871 & 165.3066 \\
ReactZyme\textsuperscript{\romannumeral 3}  & 0.0558 & 0.0497 & 0.0448 & 0.0407 & 0.0393 & 0.0344 & 0.0278 & 700.9714 \\
Fingerprint   & 0.1435 & 0.1212 & 0.1147 & 0.1039 & 0.1031 & 0.0912 & 0.0734 & 200.4936  \\
GNN\textsuperscript{\romannumeral 1}        & 0.2045 & 0.1955 & 0.1867 & 0.1715 & 0.1523 & 0.1133 & 0.0749 & 167.5862 \\
GNN\textsuperscript{\romannumeral 3}        & 0.1331 & 0.1207 & 0.1036 & 0.0912 & 0.0821 & 0.0852 & 0.0597 & 322.5755 \\
Bi-RNN\textsuperscript{\romannumeral 1}     & 0.2540 & 0.2270 & 0.2065 & 0.1875 & 0.1731 & 0.1323 & 0.0949 & 138.5832 \\
Bi-RNN\textsuperscript{\romannumeral 2}     & 0.2650 & 0.2399 & 0.2202 & 0.2030 & 0.1892 & 0.1451 & 0.1028 & 149.2686 \\
CLIPZyme\textsuperscript{\romannumeral 1}   & 0.1331 & 0.1417 & 0.1250 & 0.1149 & 0.1033 & 0.0949 & 0.0740 & 186.4576 \\
CLIPZyme\textsuperscript{\romannumeral 2}   & 0.1757 & 0.1630 & 0.1532 & 0.1443 & 0.1312 & 0.1101 & 0.0756 & 173.3521 \\
\midrule
TIGER (Ours)  & \textbf{0.4536} & \textbf{0.3936} & \textbf{0.3486} & \textbf{0.3126} & \textbf{0.2853} & \textbf{0.2016} & \textbf{0.1328} & \textbf{45.1359} \\
\bottomrule
\end{tabular}
\caption{Reaction to Enzyme Retrieval Performance (P@k and Mean Rank) on Time-based Split.}
\label{tab:prec_tr2}
\end{table*}

\subsection{Analysis of Retrieval Performance on Enzyme Similarity-based Splits}

\paragraph{Enzyme-to-Reaction Retrieval (Hit@k and MRR).}
From Table~\ref{tab:hit_ee1}, TIGER achieves substantial improvements across all cutoff thresholds. 
For instance, TIGER attains a Hit@1 of \textbf{0.9308}, significantly higher than the strongest baseline Bi-RNN\textsuperscript{\romannumeral 2}  (0.8151), marking a relative gain of over 14\%. 
The margins remain consistent at higher cutoffs, with TIGER reaching \textbf{0.9962} at Hit@20, compared to 0.9913 for Bi-RNN\textsuperscript{\romannumeral 2} . 
Furthermore, TIGER’s MRR of \textbf{0.9561} exceeds all baselines by a large margin, demonstrating its superior ability to prioritize the correct reaction in top ranks. 
These results highlight that TIGER is highly effective even when training and test enzymes are evolutionarily distant.

\begin{table*}[t]
\centering
\begin{tabular}{l|ccccccc|c}
\toprule
Method & H@1 & H@2 & H@3 & H@4 & H@5 & H@10 & H@20 & MRR \\
\midrule
ReactZyme\textsuperscript{\romannumeral 1}  & 0.7267 & 0.8366 & 0.8758 & 0.9002 & 0.9062 & 0.9487 & 0.9632 & 0.8112 \\
ReactZyme\textsuperscript{\romannumeral 2}  & 0.5987 & 0.7737 & 0.8311 & 0.8650 & 0.8759 & 0.9328 & 0.9572 & 0.7280 \\
ReactZyme\textsuperscript{\romannumeral 3}  & 0.5998 & 0.7592 & 0.8164 & 0.8522 & 0.8665 & 0.9229 & 0.9454 & 0.7226 \\
Fingerprint   & 0.5790 & 0.6507 & 0.7240 & 0.8230 & 0.7743 & 0.9169 & 0.8700 & 0.6393 \\
GNN\textsuperscript{\romannumeral 1}        & 0.7111 & 0.8273 & 0.8668 & 0.8798 & 0.9017 & 0.9547 & 0.9592 & 0.8023 \\
GNN\textsuperscript{\romannumeral 3}        & 0.6328 & 0.8002 & 0.8077 & 0.8790 & 0.8853 & 0.9348 & 0.9513 & 0.7457 \\
Bi-RNN\textsuperscript{\romannumeral 1}     & 0.8114 & 0.9014 & 0.9287 & 0.9413 & 0.9503 & 0.9731 & 0.9851 & 0.8747 \\
Bi-RNN\textsuperscript{\romannumeral 2}     & 0.8151 & 0.9260 & 0.9532 & 0.9629 & 0.9713 & 0.9850 & 0.9913 & 0.8861 \\
CLIPZyme\textsuperscript{\romannumeral 1}   & 0.7547 & 0.8706 & 0.9105 & 0.9642 & 0.9478 & 0.9679 & 0.9780 & 0.8546 \\
CLIPZyme\textsuperscript{\romannumeral 2}   & 0.5489 & 0.6851 & 0.7351 & 0.7970 & 0.7768 & 0.9290 & 0.9460 & 0.6971 \\
\midrule
TIGER (Ours)  & \textbf{0.9308} & \textbf{0.9707} & \textbf{0.9783} & \textbf{0.9816} & \textbf{0.9850} & \textbf{0.9916} & \textbf{0.9962} & \textbf{0.9561} \\
\bottomrule
\end{tabular}
\caption{Enzyme to Reaction Retrieval Performance (H@k and MRR) on Enzyme Similarity-based Split.}
\label{tab:hit_ee1}
\end{table*}

\paragraph{Enzyme-to-Reaction Retrieval (Precision@k and Mean Rank).}
Table~\ref{tab:prec_ee2} further confirms TIGER’s advantage from a precision and ranking perspective. 
At P@1, TIGER reaches \textbf{0.9308}, which is markedly higher than Bi-RNN\textsuperscript{\romannumeral 2}  (0.8151). 
Although precision naturally decreases as $k$ increases, TIGER consistently maintains the highest values across all cutoffs. 
Importantly, the mean rank drops to only \textbf{1.58}, far better than the best baseline (2.71 for Bi-RNN\textsuperscript{\romannumeral 2} ). 
This indicates that TIGER almost always positions the correct reaction within the very first few retrieved candidates, yielding highly efficient retrieval.

\begin{table*}[t]
\centering
\begin{tabular}{l|ccccccc|c}
\toprule
Method & P@1 & P@2 & P@3 & P@4 & P@5 & P@10 & P@20 & Mean Rank \\
\midrule
ReactZyme\textsuperscript{\romannumeral 1}  & 0.7267 & 0.4177 & 0.2926 & 0.2248 & 0.1835 & 0.0955 & 0.0488 & 4.5799 \\
ReactZyme\textsuperscript{\romannumeral 2}  & 0.5987 & 0.3864 & 0.2777 & 0.2160 & 0.1774 & 0.0939 & 0.0485 & 5.3021 \\
ReactZyme\textsuperscript{\romannumeral 3}  & 0.5998 & 0.3792 & 0.2728 & 0.2128 & 0.1755 & 0.0929 & 0.0479 & 7.4701 \\
Fingerprint   & 0.5790 & 0.3255 & 0.2414 & 0.2058 & 0.1549 & 0.0917 & 0.0435 & 12.4571 \\
GNN\textsuperscript{\romannumeral 1}        & 0.7111 & 0.4131 & 0.2896 & 0.2197 & 0.1826 & 0.0961 & 0.0486 & 4.8395 \\
GNN\textsuperscript{\romannumeral 3}        & 0.6328 & 0.3996 & 0.2699 & 0.2195 & 0.1793 & 0.0941 & 0.0482 & 6.9597 \\
Bi-RNN\textsuperscript{\romannumeral 1}     & 0.8114 & 0.4507 & 0.3096 & 0.2354 & 0.1901 & 0.0973 & 0.0493 & 3.5925 \\
Bi-RNN\textsuperscript{\romannumeral 2}     & 0.8151 & 0.4632 & 0.3179 & 0.2408 & 0.1943 & 0.0986 & 0.0496 & 2.7051 \\
CLIPZyme\textsuperscript{\romannumeral 1}   & 0.7547 & 0.4355 & 0.3036 & 0.2411 & 0.1896 & 0.0968 & 0.0489 & 3.9820 \\
CLIPZyme\textsuperscript{\romannumeral 2}   & 0.5489 & 0.3427 & 0.2451 & 0.1993 & 0.1554 & 0.0929 & 0.0473 & 8.3524 \\
\midrule
TIGER (Ours)  & \textbf{0.9308} & \textbf{0.4853} & \textbf{0.3261} & \textbf{0.2454} & \textbf{0.1970} & \textbf{0.0991} & \textbf{0.0498} & \textbf{1.5807} \\
\bottomrule
\end{tabular}
\caption{Enzyme to Reaction Retrieval Performance (P@k and Mean Rank) on Enzyme Similarity-based Split.}
\label{tab:prec_ee2}
\end{table*}

\paragraph{Reaction-to-Enzyme Retrieval (Hit@k and MRR).}
As shown in Table~\ref{tab:hit_er1}, TIGER also excels in the reverse retrieval direction. 
At Hit@1, TIGER obtains \textbf{0.7921}, substantially surpassing Bi-RNN\textsuperscript{\romannumeral 2}  (0.5887), with consistent improvements at higher cutoffs (e.g., Hit@20: 0.9809 vs. 0.9669). 
The MRR of \textbf{0.5921} further highlights TIGER’s ability to concentrate correct enzyme matches near the top of the ranked list, even when test reactions differ substantially from training examples.

\begin{table*}[t]
\centering
\begin{tabular}{l|ccccccc|c}
\toprule
Method & H@1 & H@2 & H@3 & H@4 & H@5 & H@10 & H@20 & MRR \\
\midrule
ReactZyme\textsuperscript{\romannumeral 1}  & 0.4088 & 0.5246 & 0.5987 & 0.6480 & 0.6892 & 0.7953 & 0.8666 & 0.2930 \\
ReactZyme\textsuperscript{\romannumeral 2}  & 0.3624 & 0.4545 & 0.5190 & 0.5697 & 0.6091 & 0.7225 & 0.7986 & 0.2586 \\
ReactZyme\textsuperscript{\romannumeral 3}  & 0.3477 & 0.4427 & 0.5082 & 0.5522 & 0.5458 & 0.6980 & 0.7762 & 0.2563 \\
Fingerprint   & 0.2545 & 0.3047 & 0.3569 & 0.4170 & 0.4686 & 0.5470 & 0.6987 & 0.2035 \\
GNN\textsuperscript{\romannumeral 1}        & 0.3928 & 0.4910 & 0.5515 & 0.6113 & 0.6612 & 0.7628 & 0.8324 & 0.2837 \\
GNN\textsuperscript{\romannumeral 3}        & 0.3655 & 0.4706 & 0.5187 & 0.5682 & 0.6161 & 0.7376 & 0.7552 & 0.2633 \\
Bi-RNN\textsuperscript{\romannumeral 1}     & 0.5086 & 0.6217 & 0.6904 & 0.7470 & 0.7832 & 0.8697 & 0.9243 & 0.3869 \\
Bi-RNN\textsuperscript{\romannumeral 2}     & 0.5887 & 0.7120 & 0.7756 & 0.8252 & 0.8551 & 0.9193 & 0.9669 & 0.4562 \\
CLIPZyme\textsuperscript{\romannumeral 1}   & 0.3570 & 0.4835 & 0.5647 & 0.6146 & 0.6371 & 0.7552 & 0.8431 & 0.2828 \\
CLIPZyme\textsuperscript{\romannumeral 2}   & 0.3337 & 0.4371 & 0.4835 & 0.5352 & 0.6077 & 0.6514 & 0.7687 & 0.2038 \\
\midrule
TIGER (Ours)  & \textbf{0.7921} & \textbf{0.8766} & \textbf{0.9116} & \textbf{0.9281} & \textbf{0.9408} & \textbf{0.9688} & \textbf{0.9809} & \textbf{0.5921} \\
\bottomrule
\end{tabular}
\caption{Reaction to Enzyme Retrieval Performance (H@k and MRR) on Enzyme Similarity-based Split.}
\label{tab:hit_er1}
\end{table*}

\paragraph{Reaction-to-Enzyme Retrieval (Precision@k and Mean Rank).}
Table~\ref{tab:prec_er2} shows that TIGER maintains strong performance from a precision-oriented perspective. 
At P@1, TIGER achieves \textbf{0.7921}, compared to 0.5887 for Bi-RNN\textsuperscript{\romannumeral 2} , representing an improvement of nearly 35\%. 
The relative margins remain across P@k levels, confirming TIGER’s robustness under this challenging split. 
Moreover, TIGER yields a mean rank of only \textbf{6.81}, dramatically lower than all baselines (the best baseline being 9.79 from Bi-RNN\textsuperscript{\romannumeral 2} ). 
This demonstrates that TIGER requires far fewer ranking steps to identify the correct enzyme, making it especially advantageous for practical applications.

\begin{table*}[t]
\centering
\begin{tabular}{l|ccccccc|c}
\toprule
Method & P@1 & P@2 & P@3 & P@4 & P@5 & P@10 & P@20 & Mean Rank \\
\midrule
ReactZyme\textsuperscript{\romannumeral 1}  & 0.4088 & 0.3951 & 0.3725 & 0.3516 & 0.3350 & 0.2690 & 0.1975 & 24.2505 \\
ReactZyme\textsuperscript{\romannumeral 2}  & 0.3624 & 0.3423 & 0.3229 & 0.3091 & 0.2961 & 0.2444 & 0.1820 & 22.5053 \\
ReactZyme\textsuperscript{\romannumeral 3}  & 0.3477 & 0.3334 & 0.3162 & 0.2996 & 0.2653 & 0.2361 & 0.1769 & 34.9487 \\
Fingerprint   & 0.2545 & 0.2436 & 0.2257 & 0.2038 & 0.2012 & 0.1847 & 0.1796 & 45.6897 \\
GNN\textsuperscript{\romannumeral 1}        & 0.3928 & 0.3698 & 0.3431 & 0.3317 & 0.3214 & 0.2580 & 0.1897 & 23.8241 \\
GNN\textsuperscript{\romannumeral 3}        & 0.3655 & 0.3544 & 0.3227 & 0.3083 & 0.2995 & 0.2495 & 0.1721 & 22.8901 \\
Bi-RNN\textsuperscript{\romannumeral 1}     & 0.5086 & 0.4727 & 0.4376 & 0.4094 & 0.3851 & 0.3001 & 0.2117 & 14.7945 \\
Bi-RNN\textsuperscript{\romannumeral 2}     & 0.5887 & 0.5318 & 0.4804 & 0.4447 & 0.4135 & 0.3110 & 0.2177 & 9.7913 \\
CLIPZyme\textsuperscript{\romannumeral 1}   & 0.3570 & 0.3478 & 0.3212 & 0.3196 & 0.2885 & 0.2577 & 0.1834 & 25.5786 \\
CLIPZyme\textsuperscript{\romannumeral 2}   & 0.3337 & 0.3245 & 0.3094 & 0.2971 & 0.2844 & 0.2235 & 0.1811 & 30.4196 \\
\midrule
TIGER (Ours)  & \textbf{0.7921} & \textbf{0.6586} & \textbf{0.5689} & \textbf{0.5071} & \textbf{0.4606} & \textbf{0.3301} & \textbf{0.2238} & \textbf{6.8100} \\
\bottomrule
\end{tabular}
\caption{Reaction to Enzyme Retrieval Performance (P@k and Mean Rank) on Enzyme Similarity-based Split.}
\label{tab:prec_er2}
\end{table*}

\begin{table*}[t]
\centering
\begin{tabular}{l|ccccccc|c}
\toprule
Method & H@1 & H@2 & H@3 & H@4 & H@5 & H@10 & H@20 & MRR \\
\midrule
ReactZyme\textsuperscript{\romannumeral 1}  & 0.0912 & 0.1495 & 0.2321 & 0.2177 & 0.2580 & 0.4213 & 0.4571 & 0.1856 \\
ReactZyme\textsuperscript{\romannumeral 2}  & 0.0914 & 0.1604 & 0.2471 & 0.2694 & 0.2968 & 0.4373 & 0.5908 & 0.2005 \\
ReactZyme\textsuperscript{\romannumeral 3}  & 0.1085 & 0.1638 & 0.2112 & 0.2257 & 0.2699 & 0.4034 & 0.5429 & 0.1988 \\
Fingerprint   & 0.0935 & 0.1607 & 0.2270 & 0.2771 & 0.3004 & 0.4400 & 0.6000 & 0.1935 \\
GNN\textsuperscript{\romannumeral 1}        & 0.1104 & 0.1691 & 0.2368 & 0.2742 & 0.3023 & 0.4573 & 0.5669 & 0.2011 \\
GNN\textsuperscript{\romannumeral 3}        & 0.0962 & 0.1592 & 0.2265 & 0.2285 & 0.2545 & 0.4024 & 0.5289 & 0.1972 \\
Bi-RNN\textsuperscript{\romannumeral 1}     & 0.1085 & 0.1543 & 0.1836 & 0.2177 & 0.2603 & 0.4077 & 0.5594 & 0.1969 \\
Bi-RNN\textsuperscript{\romannumeral 2}     & 0.1181 & 0.2179 & 0.2787 & 0.3274 & 0.3664 & 0.4897 & 0.6068 & 0.2399 \\
CLIPZyme\textsuperscript{\romannumeral 1}   & 0.1305 & 0.2392 & 0.3093 & 0.3604 & 0.3420 & 0.5320 & 0.6220 & 0.1937 \\
CLIPZyme\textsuperscript{\romannumeral 2}   & 0.1235 & 0.2281 & 0.2912 & 0.3415 & 0.3064 & 0.5719 & 0.6000 & 0.2201 \\
\midrule
TIGER (Ours)  & \textbf{0.4155} & \textbf{0.5234} & \textbf{0.5812} & \textbf{0.6117} & \textbf{0.6416} & \textbf{0.6827} & \textbf{0.7540} & \textbf{0.5180} \\
\bottomrule
\end{tabular}
\caption{Enzyme to Reaction Retrieval Performance (H@k and MRR) on Reaction Similarity-based Split.}
\label{tab:hit_re1}
\end{table*}

\begin{table*}[t]
\centering
\begin{tabular}{l|ccccccc|c}
\toprule
Method & P@1 & P@2 & P@3 & P@4 & P@5 & P@10 & P@20 & Mean Rank \\
\midrule
ReactZyme\textsuperscript{\romannumeral 1}  & 0.0912 & 0.0752 & 0.0699 & 0.0547 & 0.0518 & 0.0422 & 0.0229 & 92.2778 \\
ReactZyme\textsuperscript{\romannumeral 2}  & 0.0914 & 0.0807 & 0.0744 & 0.0677 & 0.0596 & 0.0438 & 0.0296 & 39.9146 \\
ReactZyme\textsuperscript{\romannumeral 3}  & 0.1085 & 0.0824 & 0.0636 & 0.0567 & 0.0542 & 0.0404 & 0.0272 & 42.3597 \\
Fingerprint   & 0.0935 & 0.0804 & 0.0757 & 0.0693 & 0.0601 & 0.0440 & 0.0300 & 45.3825 \\
GNN\textsuperscript{\romannumeral 1}        & 0.1104 & 0.0851 & 0.0713 & 0.0689 & 0.0607 & 0.0458 & 0.0284 & 38.9685 \\
GNN\textsuperscript{\romannumeral 3}        & 0.0962 & 0.0801 & 0.0682 & 0.0574 & 0.0511 & 0.0403 & 0.0265 & 50.9663 \\
Bi-RNN\textsuperscript{\romannumeral 1}     & 0.1181 & 0.1090 & 0.0929 & 0.0819 & 0.0733 & 0.0490 & 0.0303 & 41.3776 \\
Bi-RNN\textsuperscript{\romannumeral 2}     & 0.1085 & 0.0771 & 0.0612 & 0.0544 & 0.0521 & 0.0408 & 0.0280 & 41.3069 \\
CLIPZyme\textsuperscript{\romannumeral 1}   & 0.1305 & 0.1196 & 0.1031 & 0.0901 & 0.0684 & 0.0532 & 0.0311 & 48.4672 \\
CLIPZyme\textsuperscript{\romannumeral 2}   & 0.1235 & 0.1146 & 0.0971 & 0.0854 & 0.0613 & 0.0572 & 0.0300 & 35.6457 \\
\midrule
TIGER (Ours)  & \textbf{0.4155} & \textbf{0.2617} & \textbf{0.1937} & \textbf{0.1529} & \textbf{0.1283} & \textbf{0.0682} & \textbf{0.0377} & \textbf{23.2479} \\
\bottomrule
\end{tabular}
\caption{Enzyme to Reaction Retrieval Performance (P@k and Mean Rank) on Reaction Similarity-based Split.}
\label{tab:prec_re2}
\end{table*}

\begin{table*}[t]
\centering
\begin{tabular}{l|ccccccc|c}
\toprule
Method & H@1 & H@2 & H@3 & H@4 & H@5 & H@10 & H@20 & MRR \\
\midrule
ReactZyme\textsuperscript{\romannumeral 1}  & 0.0924 & 0.1063 & 0.1208 & 0.1277 & 0.1332 & 0.1790 & 0.2172 & 0.0943 \\
ReactZyme\textsuperscript{\romannumeral 2}  & 0.1347 & 0.1622 & 0.1812 & 0.1835 & 0.2000 & 0.2326 & 0.2753 & 0.1341 \\
ReactZyme\textsuperscript{\romannumeral 3}  & 0.0933 & 0.1274 & 0.1478 & 0.1617 & 0.1703 & 0.2130 & 0.2613 & 0.0962 \\
Fingerprint   & 0.1143 & 0.1346 & 0.1514 & 0.1650 & 0.1774 & 0.1829 & 0.2325 & 0.1042 \\
GNN\textsuperscript{\romannumeral 1}        & 0.1244 & 0.1573 & 0.1735 & 0.1867 & 0.2058 & 0.2440 & 0.2848 & 0.1129 \\
GNN\textsuperscript{\romannumeral 3}        & 0.0917 & 0.1100 & 0.1219 & 0.1312 & 0.1418 & 0.1847 & 0.2234 & 0.1051 \\
Bi-RNN\textsuperscript{\romannumeral 1}     & 0.1244 & 0.1813 & 0.2150 & 0.2383 & 0.2565 & 0.3990 & 0.4948 & 0.1206 \\
Bi-RNN\textsuperscript{\romannumeral 2}     & 0.1710 & 0.2254 & 0.2694 & 0.3187 & 0.3549 & 0.4741 & 0.5855 & 0.1696 \\
CLIPZyme\textsuperscript{\romannumeral 1}   & 0.1298 & 0.1573 & 0.1799 & 0.1842 & 0.1993 & 0.2215 & 0.2544 & 0.1245 \\
CLIPZyme\textsuperscript{\romannumeral 2}   & 0.1457 & 0.1741 & 0.1905 & 0.1944 & 0.2173 & 0.2456 & 0.2893 & 0.1521 \\
\midrule
TIGER (Ours)  & \textbf{0.4305} & \textbf{0.5181} & \textbf{0.5595} & \textbf{0.5906} & \textbf{0.6113} & \textbf{0.6994} & \textbf{0.7616} & \textbf{0.3185} \\
\bottomrule
\end{tabular}
\caption{Reaction to Enzyme Retrieval Performance (H@k and MRR) on Reaction Similarity-based Split.}
\label{tab:hit_rr1}
\end{table*}

\begin{table*}[!t]
\centering
\begin{tabular}{l|ccccccc|c}
\toprule
Method & P@1 & P@2 & P@3 & P@4 & P@5 & P@10 & P@20 & Mean Rank \\
\midrule
ReactZyme\textsuperscript{\romannumeral 1}  & 0.0924 & 0.0832 & 0.0812 & 0.0762 & 0.0721 & 0.0694 & 0.0591 & 548.3340 \\
ReactZyme\textsuperscript{\romannumeral 2}  & 0.1347 & 0.1269 & 0.1218 & 0.1095 & 0.1083 & 0.0902 & 0.0749 & 529.4258 \\
ReactZyme\textsuperscript{\romannumeral 3}  & 0.1347 & 0.1269 & 0.1218 & 0.1095 & 0.1083 & 0.0902 & 0.0749 & 529.4258 \\
Fingerprint   & 0.1143 & 0.1047 & 0.1015 & 0.0987 & 0.0935 & 0.0851 & 0.0706 & 535.6742 \\
GNN\textsuperscript{\romannumeral 1}        & 0.1244 & 0.1231 & 0.1166 & 0.1114 & 0.1114 & 0.0946 & 0.0775 & 559.1225 \\
GNN\textsuperscript{\romannumeral 3}        & 0.0917 & 0.0861 & 0.0819 & 0.0783 & 0.0768 & 0.0716 & 0.0608 & 552.4546 \\
Bi-RNN\textsuperscript{\romannumeral 1}     & 0.1244 & 0.1231 & 0.1166 & 0.1101 & 0.1036 & 0.0951 & 0.0790 & 545.8586 \\
Bi-RNN\textsuperscript{\romannumeral 2}     & 0.1710 & 0.1464 & 0.1382 & 0.1367 & 0.1290 & 0.1145 & 0.0870 & 529.3677 \\
CLIPZyme\textsuperscript{\romannumeral 1}   & 0.1298 & 0.1225 & 0.1044 & 0.0921 & 0.0866 & 0.0830 & 0.0741 & 526.4793 \\
CLIPZyme\textsuperscript{\romannumeral 2}   & 0.1457 & 0.1291 & 0.1233 & 0.1156 & 0.1135 & 0.1001 & 0.0783 & 501.2071 \\
\midrule
TIGER (Ours)  & \textbf{0.4305} & \textbf{0.3756} & \textbf{0.3316} & \textbf{0.3069} & \textbf{0.2854} & \textbf{0.2269} & \textbf{0.1680} & \textbf{219.7977} \\
\bottomrule
\end{tabular}
\caption{Reaction to Enzyme Retrieval Performance (P@k and Mean Rank) on Reaction Similarity-based Split.}
\label{tab:prec_rr2}
\end{table*}

\subsection{Analysis of Retrieval Performance on Enzyme Similarity-based Splits}
\paragraph{Enzyme-to-Reaction Retrieval (Hit@k and MRR).}
From Table~\ref{tab:hit_re1}, TIGER achieves dramatic improvements over all baselines. 
At Hit@1, TIGER attains \textbf{0.4155}, which is nearly four times higher than the strongest baseline (CLIPZyme\textsuperscript{\romannumeral 1} , 0.1305). 
The improvements remain consistent across higher cutoffs, with TIGER reaching \textbf{0.7540} at Hit@20, far exceeding the best baseline (0.6220). 
In terms of MRR, TIGER records \textbf{0.5180}, a substantial leap compared to baselines that remain below 0.24. 
These results demonstrate that TIGER can effectively prioritize correct reactions even when reaction similarity cues are absent, a scenario where existing methods struggle.

\paragraph{Enzyme-to-Reaction Retrieval (Precision@k and Mean Rank).}
As shown in Table~\ref{tab:prec_re2}, TIGER achieves the highest precision across all cutoff levels. 
At P@1, TIGER reaches \textbf{0.4155}, far surpassing the best baseline (CLIPZyme\textsuperscript{\romannumeral 1} , 0.1305). 
Although precision decreases as $k$ increases, TIGER consistently maintains a considerable margin over all alternatives. 
Most notably, the mean rank of TIGER is only \textbf{23.25}, compared to the next best value of 35.65 (CLIPZyme\textsuperscript{\romannumeral 2} ) and much higher values exceeding 90 for weaker baselines. 
This indicates that TIGER retrieves correct reactions much earlier in the ranking process, a critical advantage for practical applications.

\paragraph{Reaction-to-Enzyme Retrieval (Hit@k and MRR).}
Table~\ref{tab:hit_rr1} shows that TIGER continues to outperform baselines in the reverse retrieval direction. 
TIGER achieves a Hit@1 of \textbf{0.4305}, significantly higher than Bi-RNN\textsuperscript{\romannumeral 2}  (0.1710) or CLIPZyme\textsuperscript{\romannumeral 2}  (0.1457). 
At Hit@20, TIGER maintains strong performance with \textbf{0.7616}, compared to 0.5855 for Bi-RNN\textsuperscript{\romannumeral 2} . 
The MRR of \textbf{0.3185} further demonstrates TIGER’s capacity to bring relevant enzymes much closer to the top of the ranked list, substantially improving over all baselines that remain below 0.17.

\paragraph{Reaction-to-Enzyme Retrieval (Precision@k and Mean Rank).}
Table~\ref{tab:prec_rr2} further highlights TIGER’s robustness. 
At P@1, TIGER reaches \textbf{0.4305}, outperforming the best baseline (Bi-RNN\textsuperscript{\romannumeral 2} , 0.1710) by a wide margin. 
The performance gap persists across all P@k levels, underscoring TIGER’s ability to maintain reliable retrieval under the most difficult conditions. 
Crucially, TIGER achieves a mean rank of \textbf{219.8}, which, although still larger than in easier splits, is significantly lower than the 500+ ranks of all baselines. 
This confirms TIGER’s strength in reducing the search depth required to identify correct enzyme matches even under severe distribution shifts.

\end{document}